%% file: thesis.tex
\documentclass[
  oneside
]{book}
\pdfoutput=1
\usepackage[T1]{fontenc}
\usepackage{amsmath,amssymb,amsthm,amsfonts}
\usepackage[pdftex]{graphicx}
\usepackage{caption}
\usepackage{subcaption}
\usepackage[
    oldstylenums,
    largesmallcaps,
]{kpfonts}
\usepackage[colorinlistoftodos]{todonotes}
\usepackage{hyperref}
\hypersetup{
    colorlinks,
    breaklinks,
    pdftitle  = {On the Linear Algebraic Structure of Distributed Word Representations},
    pdfauthor = {Lisa Seung-Yeon Lee}
}
\usepackage{geometry}
\usepackage{lastpage}
\usepackage{fancyhdr}
\makeatletter
\renewcommand\chapter{\if@openright\cleardoublepage\else\clearpage\fi
                    \thispagestyle{empty}% original style: plain
                    \global\@topnum\z@
                    \@afterindentfalse
                    \secdef\@chapter\@schapter}
\makeatother
\renewcommand{\chaptermark}[1]{
\markboth{\chaptername\
\thechapter.\ #1}{}}
\usepackage{mathtools} %\DeclarePairedDelimiter
\usepackage{color} %\hilight
\usepackage{array}
\newcolumntype{t}{>{\ttfamily}l}
\usepackage{enumerate}
\usepackage{tikz}
\usetikzlibrary{positioning,chains,fit,shapes,calc}

\usepackage{fancyvrb} %\VerbatimInput
\usepackage{framed}
\usepackage{lipsum}
\newcommand\abstractname{Abstract}  %%% here
\makeatletter
\if@titlepage
  \newenvironment{abstract}{%
      \titlepage
      \null\vfil
      \@beginparpenalty\@lowpenalty
      \begin{center}%
        \bfseries \abstractname
        \@endparpenalty\@M
      \end{center}}%
     {\par\vfil\null\endtitlepage}
\else
  \newenvironment{abstract}{%
      \if@twocolumn
        \section*{\abstractname}%
      \else
        \small
        \begin{center}%
          {\bfseries \abstractname\vspace{-.5em}\vspace{\z@}}%
        \end{center}%
        \quotation
      \fi}
      {\if@twocolumn\else\endquotation\fi}
\fi
\makeatother

\theoremstyle{remark}

\newcommand{\eat}[1]{}

\newcommand{\hilight}[1]{\colorbox{yellow}{#1}}

\DeclarePairedDelimiter{\paren}{(}{)}

\DeclarePairedDelimiter{\len}{\|}{\|}
\DeclareMathOperator*{\argmin}{arg\,min}
\DeclareMathOperator*{\argmax}{arg\,max}

\DeclareMathOperator*{\KB}{KB}
\DeclareMathOperator*{\acc}{acc}
\DeclareMathOperator*{\pos}{pos}
\DeclareMathOperator*{\lex}{lex}
\DeclareMathOperator*{\PMI}{PMI}

\DeclareMathOperator{\E}{\mathbb{E}}

\newcommand*{\Cdot}{\raisebox{-0.25ex}{\scalebox{1.2}{$\cdot$}}}

\begin{document}
%%%%%%%%%%%%%%%%%%%%%%%%%%%%%%%%%%%%%%%%%%%%%%%%%%%%%%%%%%%%%%%%%%%%%%%%%%%%%%%%

	\pagestyle{empty}
	\title{On the Linear Algebraic Structure of \\Distributed Word Representations}
	\author{Lisa Seung-Yeon Lee\\\\Advised by Professor Sanjeev Arora}
	%\publisher{Princeton University Department of Mathematics}
	%\end title
	\date{4 May 2015\vskip 4in A thesis submitted to the Princeton University Department of Mathematics\\ in partial fulfillment of the requirements for the degree of Bachelor of Arts.}
	\maketitle
	
\begin{abstract}
In this work, we leverage the linear algebraic structure of distributed word representations to automatically extend knowledge bases and allow a machine to learn new facts about the world. Our goal is to extract structured facts from corpora in a simpler manner, without applying classifiers or patterns, and using only the co-occurrence statistics of words. We demonstrate that the linear algebraic structure of word embeddings can be used to reduce data requirements for methods of learning facts. In particular, we demonstrate that words belonging to a common category, or pairs of words satisfying a certain relation, form a low-rank subspace in the projected space. We compute a basis for this low-rank subspace using singular value decomposition (SVD), then use this basis to discover new facts and to fit vectors for less frequent words which we do not yet have vectors for.
\end{abstract}
  %\newpage
This thesis represents my own work in accordance with university regulations.
\vskip .25in

\hskip 3in \includegraphics[width=0.3\textwidth]{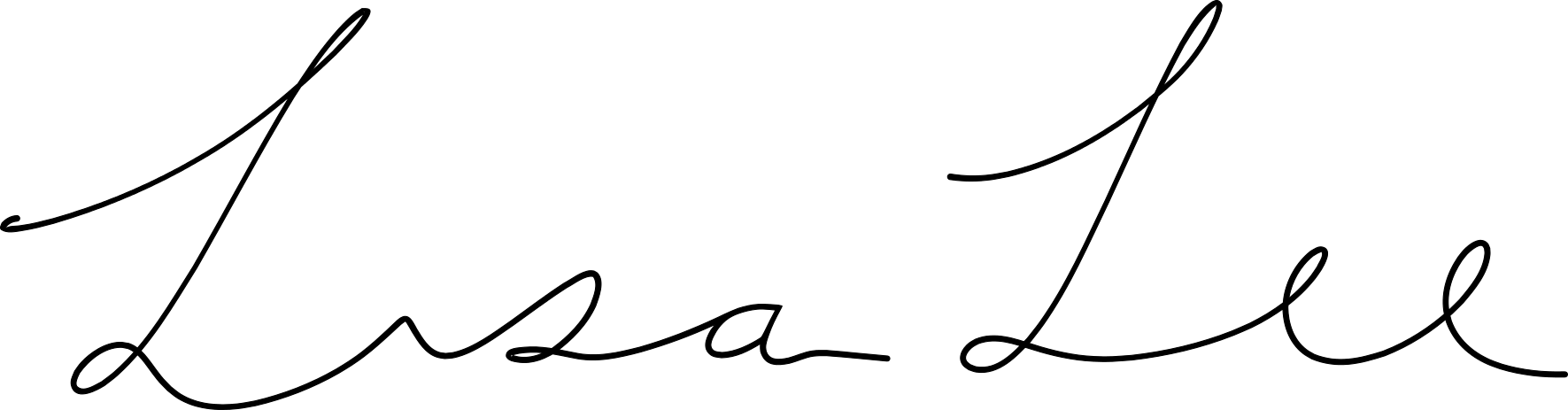}

\vskip .125in
\hskip 4.15in Lisa Lee

	%% The front matter contains tables, epigraphs, the foreword, etc.
	\frontmatter

	\tableofcontents
        %\listoffigures
        %\listoftables
	\chapter*{Acknowledgements}

First and foremost, I would like to thank Professor Sanjeev Arora for his patient guidance, advice, and support during the planning and development of this research work. I also would like to express my very great appreciation to Dr. Yingyu Liang, without whom I could not have completed this research project; thank you so much for the time and effort you took to help me run experiments and understand concepts, and for suggesting new ideas and directions to take when I felt stuck. I would also like to deeply thank Tengyu Ma for his valuable ideas and suggestions during this research project.

I am also greatly indebted to Ming-Yee Tsang, who helped me format my thesis, and debug the complicated regular expressions code for preprocessing the huge Wikipedia corpus (see Section \ref{section:preprocessing_corpus}). Many thanks to Victor Luu, Irene Lo, Eliott Joo,  Christina Funk, and Ante Qu for proofreading my thesis and providing invaluable feedback.

Channing, thank you so much for keeping me company while I was writing this thesis, for impromptu coffee runs in the middle of the night so that I wouldn't fall asleep, for always motivating and encouraging me, and a myriad other things.

I would also like to thank my wonderful friends and famiLee for their support. Mimi, thank you for being the silliest, kindest, coolest, most patient sister and best friend that you are to me. Joonhee, you will always be my cute little brother no matter how old you are. Thank you for all the fun Naruto missions we went on when you were little, for all your hilarious jokes, and for playing LoL/Pokemon/Minecraft with me. Happy, I love you too. Too bad you can't read this. Woof woof. Thank you Laon for always being by my side since I was five. Thank you umma and abba for raising me and my siblings (and Happy), and for always encouraging me to try my best. 

Billy, thanks for always challenging me to think more rigorously, and for all the fun music-making we did (yay Brahms, Rachmaninoff, Saint-Saens, Chopin, Piazzolla, Hisaishi, Pokemon). I am also extremely grateful to have found such a loving community in Manna Christian Fellowship during my freshman year. 

Last but not least, thank you God for giving me these four precious years at Princeton University to meet all these wonderful people and to study math.
	
	%% The main matter (normal chapters)
	\pagestyle{fancy}
	\setcounter{page}{0}
	\mainmatter
        \input{intro.tex}

        \input{background.tex}

        \input{methods.tex}
        \input{category_relations.tex}
        \input{capture_rate.tex}
        \input{learn_new_facts.tex}
        \input{fit_new_vec.tex}

        \input{wordnet_filter.tex}
        \input{conclusion.tex}

	%% The back matter contains the appendix, bibliography, index, glossary, etc.
	\backmatter
	\appendix
	\bibliographystyle{plain}
	\bibliography{thesis}
\end{document}

%% file: intro.tex
\chapter{Introduction}

\section{Distributed word representations}

Distributed representations of words in a vector space represent each word with a real-valued vector, called a \emph{word vector}. They are also known as \emph{word embeddings} because they embed an entire vocabulary into a relatively low-dimensional linear space whose dimensions are latent continuous features. One of the earliest ideas of distributed representations dates back to 1986 \cite{hinton1986learning}, and has since been applied to statistical language modeling with considerable success. These word vectors have shown to improve performance in a variety of natural language processing tasks including automatic speech recognition \cite{Schwenk:2007:CSL:1230156.1230409}, information retrieval \cite{manning2008}, document classification \cite{Sebastiani02machinelearning}, and parsing \cite{SocherEtAl2013:CVG}.

The word vectors are trained over large corpora\footnote{A \emph{corpus} (plural \emph{corpora}) is a large and structured set of unlabeled texts.} typically in a totally unsupervised manner, using the co-occurrence statistics of words.\footnote{We say two words \emph{co-occur} in a corpus if they appear together within a certain (fixed) distance in the text.} Past methods to obtain word embeddings include matrix factorization methods \cite{deerwester-indexing-1990}, variants of neural networks \cite{Rumelhart:1988:LRB:65669.104451, nnlm:2001:nips, Bengio03aneural, collobert08, NIPS2008_3583, mikolov2011, MSC+13}, and energy-based models \cite{GloVe, arora}. The learned word vectors explicitly capture many linguistic regularities and patterns, such as semantic and syntactic attributes of words. Therefore, words that appear in similar contexts, or belong to a common ``category'' (e.g., country names, composer names, or university names), tend to form a cluster in the projected space.

Recently, Mikolov et al.\ \cite{MSC+13} demonstrated that word embeddings created by a recurrent neural net (RNN) and by a related energy-based model called \texttt{word2vec} exhibit an additional \emph{linear} structure which captures the relation between \emph{pairs} of words, and allows one to solve analogy queries such as ``man:woman::king:??'' using simple vector arithmetics. More specifically, ``queen'' happens to be the word whose vector $v_{\textrm{queen}}$ is the closest approximation to the vector $v_{\textrm{woman}} - v_{\textrm{man}} + v_{\textrm{king}}$. Other subsequent works \cite{GloVe, LG14a, NIPS2013_5165} produced word vectors that can be used to solve analogy queries in the same way. It remains a mystery as to why these radically different embedding methods, including highly non-linear ones, produce vectors exhibiting similar linear structure. A summary of current justifications for this phenomenon is provided in Section \ref{section:justification}.

\section{Extending existing knowledge bases}

In this work, we aim to leverage the linear algebraic structure of word embeddings to extend \emph{knowledge bases}\footnote{A \emph{knowledge base} is a collection of information that represents facts about the world.} and learn new facts. Knowledge bases such as Wordnet \cite{wn} or Freebase \cite{freebase} are a key source for providing structured information about general human knowledge. Building such knowledge bases, however, is an extremely slow and labor-intensive process. Consequently, there has been much interest in finding methods for automatically learning new facts and extending knowledge bases, e.g., by applying patterns or classifiers on large corpora \cite{snow2005learning, fader2011identifying, suchanek2007ycs}. Carlson et al.'s NELL (Never-Ending Language Learning) system \cite{Carlson10towardan}, for instance, extracts structured facts from the web to build a knowledge base, using over 1500 different classifiers and extraction methods in combination with a large-scale semi-supervised multi-task learning algorithm.

Our goal is to extract structured facts from corpora in a \emph{simpler} manner, without applying classifiers or patterns, and using only the co-occurrence statistics of words. More specifically, we use the word co-occurrence statistics to produce word vectors, and then leverage their linear structure to learn new facts, such as new words belonging to a known \emph{category}, or new pairs of words satisfying a known \emph{relation} (see Chapter \ref{chapter:category_relations}). Our methods can supplement other methods for extending knowledge bases to reduce false positive rate, or narrow down the search space for discovering new facts.

\section{Overview of the paper}

In this paper, we will demonstrate that the linear algebraic structure of word embeddings can be used to reduce data requirements for methods of learning facts. In Chapter \ref{chapter:background}, we present a few methods for learning word vectors, and provide intuition as to why the embedding methods work. Chapter \ref{chapter:methods} describes how the word vectors used in our experiments were trained. Chapter \ref{chapter:category_relations} introduces the notion of \emph{categories} and \emph{relations}, which can be used to represent facts about the world in a knowledge base. In Chapter \ref{chapter:capture_rate}, we explore the linear algebraic structure of word embeddings. In particular, we demonstrate that words belonging to a common category, or pairs of words satisfying a certain relation, form a low-rank subspace in the projected space. We compute a basis for this low-rank subspace using singular value decomposition (SVD), then use this basis to discover new facts (Chapter \ref{chapter:learn_new_facts}) and to fit vectors for less frequent words which we do not yet have vectors for (Chapter \ref{chapter:fit_new_vec}). We also demonstrate that, using an external knowledge source such as Wordnet \cite{wn}, one can improve accuracy on analogy queries of the form ``a:b::c:??'' (Chapter \ref{chapter:wordnet_filter}).

\eat{\hilight{(1990s: SVD-only method, which was distance-based, not linear algebraic)}}

%% file: background.tex
\chapter{Word embeddings}\label{chapter:background}

In this chapter, we introduce the reader to recent word embedding methods, and provide justifications for why these methods work.

In Section \ref{section:word_vector_methods}, we present three different methods for producing word vectors that exhibit the desired linear properties\footnote{That is, allowing one to solve analogy queries using linear algebraic vector arithmetics.}: Mikolov et al.'s skip-gram with negative sampling (SGNS) method \cite{MSC+13}, Pennington et al.'s GloVe method \cite{GloVe}, and Arora et al.'s Squared Norm (SN) objective \cite{arora}. In Section \ref{section:justification}, we provide a summary of current justifications for why these methods work. For further details and evaluations of these methods, see \cite{MSC+13, GloVe, arora}.

The three methods presented in Section \ref{section:word_vector_methods} achieve similar, state-of-the-art performance on analogy query tasks. In our experiments, we use the SN objective (\ref{eq:objective-sq}) to train the word vectors because it is perhaps the simplest method thus far for fitting word embeddings, and it is also the only method out of the three which \emph{provably} finds the near-optimum fit\footnote{(according to their generative model for text corpora described in their paper \cite{arora})} (see Arora et al.\ \cite{arora}).

\section{Notation}

We first introduce some notation. Let $\mathcal{D}$ be the set of distinct words that appear in a corpus $\mathcal{C}$; then we say $\mathcal{D}$ is a set of \emph{vocabulary} words that appear in $\mathcal{C}$. We can enumerate the sequence of words (or \emph{tokens}) in $\mathcal{C}$ as $w_1, w_2, \ldots, w_{|\mathcal{C}|}$, where $|\mathcal{C}|$ is the total number of tokens in $\mathcal{C}$. Let $k \in \mathbb{N}$ be fixed; then the \emph{context window} of size $k$ around a word $w_i \in \mathcal{C}$ is the multiset consisting of the $k$ tokens appearing before and after $w_i$ in the corpus,
\[
\textrm{window}_k(w_i) := \{ w_{i-k}, w_{i-k+1}, \ldots, w_{i-1}\} \cup \{w_{i+1}, w_{i+2}, \ldots, w_{i+k} \}.
\]
Typically, the context window size $k$ is chosen to be a fixed number between $5$ and $10$. For a vocabulary word $w \in \mathcal{D}$, let
\[
\kappa(w) := \bigcup_{\substack{i \in \{1, \ldots, |\mathcal{C}|\}:\\ w_i = w}} \textrm{window}_k(w_i)
\]
be the set of all tokens appearing in some context window around $w$. Each distinct word $\chi \in \kappa(w)$ is called a \emph{context} word for $w$.\footnote{This is just one way of defining \emph{context}, and other types of contexts can be considered; see \cite{levy2014dependencybased}.} Let $\mathcal{D}_{\textrm{context}}$ be the set of all context words; that is, $\mathcal{D}_{\textrm{context}}$ is the set of all distinct words in $\cup_{w \in \mathcal{D}} \kappa(w)$. Note that, because of the way we define context, we have $\mathcal{D}_{\textrm{context}} = \mathcal{D}$ and the distinction between a word and a context word is arbitrary, i.e.,  we are free to interchange the two roles.

For two words $w, w' \in \mathcal{D}$, let
\[
X_{ww'} := |\{ w_i \in \kappa(w) : w_i = w' \}|
\]
i.e., $X_{ww'}$ is the number of times word $w'$ appears in any context window around $w$. Then $X_w := \sum_{w'} X_{ww'} = |\kappa(w)|$, and $p(w' \mid w) := {X_{ww'} \over X_w}$ is the empirical probability that word $w'$ appears in \emph{some} context window around $w$ (i.e., $w'$ is a context word for $w$). Also, $p(w) := \frac{X_w}{\sum_{w'} X_{w'}}$ is the empirical probability that a randomly selected word of the corpus is $w$. The matrix $X$ whose rows and columns are indexed by the words in $\mathcal{D}$, and whose entries are $X_{ww'}$, is called the \emph{word co-occurrence matrix} of $\mathcal{C}$.

In this paper, let $d \in \mathbb{N}$ be the dimension\footnote{The dimension $d$ of the word vectors is a parameter that can be chosen, and is typically much smaller than the number of vocabulary words $|\mathcal{D}|$ or the size of the corpus $|\mathcal{C}|$. In the three methods presented in Section \ref{section:word_vector_methods}, a dimension between $50$ and $300$ is used.}  of the word vectors $v_w \in \mathbb{R}^d$. The word co-occurrence statistics are used to train the word vectors $v_w \in \mathbb{R}^d$ for words $w \in \mathcal{D}$.

%%%%%%%%%%%%%%%%%%%%%%%%%%%%%%%%%%%%%%%%%%%%%%%%%%%%%%%%%%%%%%%%%%%%%%%%%%%%%%%
\section{Methods for learning word vectors}\label{section:word_vector_methods}

Below, we present a few methods for obtaining word embeddings which allow one to solve analogy queries using linear algebraic vector arithmetics. Other methods include large-dimensional embeddings that explicitly encode co-occurrence statistics \cite{LG14a} (see Section \ref{section:justification}) and noise-contrastive estimation \cite{NIPS2013_5165}.

\subsection{Skip-gram}

For a word $w \in \mathcal{D}$ and a context $\chi \in \mathcal{D}_{\textrm{context}}$, we say the pair $(w,\chi)$ is \emph{observed in the corpus} and write $(w,\chi) \in \mathcal{C}$, if $\chi$ appears in some context window around $w$ (i.e., $\chi \in \kappa(w)$). In Mikolov et al.'s skip-gram with negative sampling (SGNG) model \cite{MSC+13, GL14}, the probability that a word-context pair $(w,\chi)$ is observed in the corpus is parametrized by
\begin{equation*}
p( w, \chi) = \frac{1}{1+\exp\paren*{-v_w \cdot v_{\chi}}}\,,
\end{equation*}
where $v_w, v_{\chi} \in \mathbb{R}^d$ are the vectors for $w$ and $\chi$ respectively. SGNS tries to maximize $p(w, \chi)$ for observed $(w,\chi)$ pairs in the corpus $\mathcal{C}$, while minimizing $p( w, \chi)$ for randomly sampled ``negative'' samples\footnote{We assume that the randomly generated negative samples $(w,\chi)$ are not observed in the corpus.} $(w,\chi)$. Their optimization objective is the log-likelihood,
\begin{align*}
&\argmax_{\{ v_w : w \in \mathcal{D} \}} \;\paren*{ \prod_{(w,\chi) \in \mathcal{C}} p(w, \chi)} \paren*{ \prod_{(w,\chi) \notin \mathcal{C}} \paren*{1 - p( w, \chi)} }\\
=
&\argmax_{\{ v_w : w \in \mathcal{D} \}} \sum_{(w,\chi) \in \mathcal{C}} \log p( w, \chi) + \sum_{(w,\chi) \notin \mathcal{C}} \log \paren*{1 - p(w, \chi)} \,,
\end{align*}
where $\argmax_x f(x) := \{ x : f(x) \geq f(y) \;\forall y \}$ is the set of arguments for which the given function $f$ attains its maximum value.

\subsection{GloVe}

Pennington et al.'s GloVe (``Global Vectors for Word Representation'') method \cite{GloVe} fits, for each word $w \in \mathcal{D}$, two low-dimensional vectors $v_w, \tilde{v}_w \in \mathbb{R}^d$ and scalars $b_w, \tilde{b}_w \in \mathbb{R}$ so as to minimize the cost function
\begin{equation}
\sum_{w,w'} f(X_{ww'}) \paren*{ v_w \cdot \tilde{v}_{w'} + b_w + \tilde{b}_{w'} - \log X_{ww'} }^2\,,
\label{eq:objective-glove}
\end{equation}
where $f(x) = \min\left\{ \paren*{x \over x_{\max}}^\alpha, 1 \right\}$ for some constants $x_{\max}$ and $\alpha$. In their experiments, they use $\alpha = 0.75$ and $x_{\max} = 100$. The purpose of the weighing function $f$ is twofold:
\begin{itemize}
\item $f(x)$ is non-decreasing, so that rare co-occurrences (which tend to have greater noise) are not overweighted.
\item $f(x)$ is relatively small for large values of $x$, so that frequent co-occurrences are not overweighted.
\end{itemize}

For the motivation behind the GloVe objective (\ref{eq:objective-glove}), we refer the reader to their paper \cite{GloVe}.

\subsection{Squared Norm}

Arora et al.'s Squared Norm (SN) method \cite{arora} fits a scalar $Z \in \mathbb{R}$ and, for each word $w \in \mathcal{D}$, a  vector $ v_w \in \mathbb{R}^d$ so as to minimize the objective
\begin{equation}
\sum_{w, w'} f(X_{ww'}) \paren*{ \log(X_{w,w'}) - \| v_w + v_{w'}\|^2 - Z }^2\,,
\label{eq:objective-sq}
\end{equation}
where\footnote{In this paper, $\len{\Cdot}$ refers to Euclidean norm, or $\len{\Cdot}_2$.} $f$ is the same weighting function as in the GloVe objective (\ref{eq:objective-glove}). For the motivation behind the SN objective (\ref{eq:objective-sq}), see Section \ref{section:sq-motivation}.

%%%%%%%%%%%%%%%%%%%%%%%%%%%%%%%%%%%%%%%%%%%%%%%%%%%%%%%%%%%%%%%%%%%%%%%%%%%%%%%
 
\section{Justification for why the word embedding methods work}\label{section:justification}

Once the word vectors have been produced, one can answer analogy queries of the form ``a:b::c:??'' by finding the word $\tilde{d}$ whose vector is closest to $v_b - v_a + v_c$ according to cosine similarity, i.e.,
\begin{align}
\tilde{d}
&= \argmax_{d} v_d \cdot (v_b - v_a + v_c)\nonumber\\
&= \argmin_d (v_a - v_b - v_c) \cdot v_d\nonumber\\
&= \argmin_d 2(v_a - v_b - v_c) \cdot v_d + \| v_a - v_b - v_c \|^2 + \|v_d\|^2\nonumber\\
&= \argmin_d \| v_a - v_b - v_c + v_d \|^2 \label{eq:analogy-query-mod}
\end{align}
where\footnote{The second to last equality follows because $\|v_d\|^2 = 1$ is a constant.} the vectors $v_w$ for each word $w$ have been normalized so that $\len{v_w} = 1$.

It remains a mystery as to why such vastly different embedding methods, including highly nonlinear ones, produce vectors exhibiting similar linear structure, and achieve fairly similar accuracy on analogy queries. In the rest of this chapter, we provide a summary of the current justifications for this phenomenon.

\subsection{Justification for explicit, high-dimensional word embeddings}
Levy and Goldberg \cite{LG14a} and Pennington et al.\ \cite{GloVe} provide a statistical intuition as to why the answer to the analogy ``man:woman::king:??'' must be ``queen''. The reason is that most contexts $\chi \in \mathcal{D}_{\textrm{context}}$ satisfy
\[
\frac{p(\chi \mid \textrm{man})}{p(\chi \mid \textrm{woman})} \approx \frac{p(\chi \mid \textrm{king})}{p(\chi \mid \textrm{queen})}
\]
where $p(\chi \mid w)$ is the conditional probability that $\chi$ appears in some context window around word $w$. For example, both ratios are around 1 for most contexts (e.g., ``sleep'', ``the'', ``food''), but the ratio deviates from 1 when $\chi$ is not gender-neutral (e.g., ``dress'', ``he'', ``she'', ``Elizabeth'', ``Henry''). Therefore, a reasonable answer to the analogy query ``a:b::c:??'' is the word $d$ that minimizes
\begin{equation}
\sum_{\chi \in \mathcal{D}_{\textrm{context}}} \paren*{ \log \frac{p(\chi \mid a)}{p(\chi \mid b)} - \log \frac{p(\chi \mid c)}{p(\chi \mid d)}  }^2.
\label{eq:LG14a-objective}
\end{equation}
Hence, Levy and Goldberg \cite{LG14a} proposed a very high-dimensional embedding that explicitly encodes correlation statistics between words and contexts: The vector for word $w \in \mathcal{D}$ is given by $v_w \in \mathbb{R}^{|\mathcal{D}_{\textrm{context}}|}$, where $v_w$ is indexed by all possible contexts $\chi \in \mathcal{D}_{\textrm{context}}$, and the entry $(v_w)_\chi$ in coordinate $\chi$ is equal to
\begin{equation}
\PMI(w, \chi) := \log \frac{p(w,\chi)}{p(w) p(\chi)} = \log \frac{p(\chi \mid w)}{p(\chi)}\,.
\label{eq:pmi}
\end{equation}
Note that with this word embedding, the objective (\ref{eq:LG14a-objective}) is \emph{equivalent} to (\ref{eq:analogy-query-mod}):
\begin{align*}
\min_d \| v_a - v_b - v_c + v_d\|_2^2
&= \min_d \sum_{\chi} \paren*{ \log \frac{p(\chi \mid a)}{p(\chi)} - \log \frac{p(\chi \mid b)}{p(\chi)} - \log \frac{p(\chi \mid c)}{p(\chi)} + \log \frac{p(\chi \mid d)}{p(\chi)} }^2\\
&= \min_d \sum_\chi \paren*{ \log {p(\chi \mid a) \over p(\chi \mid b) } - \log{p(\chi \mid c) \over p(\chi \mid d)} }^2.
\end{align*}
And indeed, Levy and Goldberg \cite{LG14a} show that the explicit word embeddings solve analogies via linear algebraic queries empirically.

\subsection{Justification for low-dimensional embeddings} \label{section:justification-arora}
However, the above justification only applies to \emph{large}-dimensional embeddings that explicitly encode correlation statistics between words and contexts. Recently, Arora et al.\ \cite{arora} gave a justification for why \emph{low}-dimensional embeddings are successful in solving analogy queries. More specifically, they postulate that  the PMI matrix\footnote{The PMI (pointwise mutual information) matrix is a $|\mathcal{D}| \times |\mathcal{D}_{\textrm{context}}|$ matrix whose rows are indexed by words $w \in \mathcal{D}$ and columns are indexed by contexts $\chi \in \mathcal{D}_{\textrm{context}}$, and whose entries are $\PMI(w,\chi)$ as defined in (\ref{eq:pmi}).} and the word vectors $v_w \in \mathbb{R}^d$ satisfy the following properties:

\begin{enumerate}[Property A.]
\item The PMI matrix can be approximated by a positive semidefinite matrix of fairly low rank, which is closer to $\log n$ than to $n$. This yields natural word embeddings $v_w$ that are implicit in the co-occurrence data itself: $\PMI(w,w') \approx v_w \cdot v_{w'}$.
\item The word vectors $v_w$ are approximately \emph{isotropic} meaning the expectation $\E_w[v_w v_w^T]$ is approximately like the identity matrix $I$, in that every one of its eigenvalues lies in $[1,1+\delta]$ for some small $\delta > 0$.
\end{enumerate}

They provide a plausible \emph{generative model} for text corpora using log-linear distributions\footnote{The generative model directly models how words are produced as the corpus is being generated, and captures latent semantic structure in the text. For details about their generative model, see \cite{arora}. For an overview of log-linear models, which are very widely used in natural language processing, see \cite{loglinearmodels}.}, under which both properties hold. Using this generative model, they prove that, up to a constant $\log Z$ and some small error,
\begin{align}
\log p(w,w') &\propto \len{v_w + v_{w'}}^2 - 2 \log Z \label{eq:arora-theorem1}\\
\log p(w) &\propto \len{v_w}^2 - \log Z \label{eq:arora-theorem2}
\end{align}
where $p(w,w')$ is the probability that the words $w$ and $w'$ appear together in the corpus, and $p(w)$ is the prior of seeing word $w$ in the corpus. Therefore,
\begin{align}
\PMI(w,w')
&:= \log p(w,\chi) - \log p(w) - \log p(\chi) \nonumber\\
&\propto \len{v_w + v_{w'}}^2 -  \len{v_w}^2 -  \len{v_{w'}}^2 \quad\qquad\textrm{by (\ref{eq:arora-theorem1}) and (\ref{eq:arora-theorem2})} \nonumber\\
&=  2 v_w \cdot v_{w'}\nonumber\\
&\propto v_w \cdot v_{w'} \,. \label{eq:arora-justification}
\end{align}

Property B implies that, for every vector $v \in \mathbb{R}^d$,
\begin{align*}
\len{v}^2
&\approx v^T \E_w[v_w v_w^T] v \qquad\textrm{(up to error $1+\delta$)}\\
&= \E_w[(v \cdot v_w)^2]\,,
\end{align*}
and so the query (\ref{eq:analogy-query-mod}) for solving ``a:b::c:??'' becomes
\begin{align*}
\argmin_d \| v_a - v_b - v_c + v_d \|^2 
&\approx \argmin_d \E_w \paren*{ v_a \cdot v_w - v_b \cdot  v_w - v_c \cdot v_w + v_d \cdot v_w }^2\\
&= \argmin_d \E_w\paren*{ \PMI(a,w) - \PMI(b,w) - \PMI(c,w) + \PMI(d,w)  }^2 \qquad\textrm{by (\ref{eq:arora-justification})}\\
&= \argmin_d \E_w\paren*{ \log \frac{p(\chi \mid a)}{p(\chi)} - \log \frac{p(\chi \mid b)}{p(\chi)} - \log \frac{p(\chi \mid c)}{p(\chi)} + \log \frac{p(\chi \mid d)}{p(\chi)} }^2\\
&= \argmin_d \E_w\paren*{ \log {p(w \mid a) \over p(w \mid b) } - \log{p(w \mid c) \over p(w \mid d)}}^2\,,
\end{align*}
which is close to (\ref{eq:LG14a-objective}), with word $w$ acting as context $\chi$.

%%%%%%%%%%%%%%%%%%%%%%%%%%%%%%%%%%%%%%%%%%%%%%%%%%%%%%%%%%%%%%%%%%%%%%%%%%%%%%
%
\subsection{Motivation behind the Squared Norm objective}\label{section:sq-motivation}
Note that by (\ref{eq:arora-theorem1}), we have $\log p(w, w') \propto \| v_w + v_{w'} \|_2^2 + C$ where $C := - 2 \log Z$ is an unknown constant. Then since the empirical probability $p(w,w')$ is given by
\begin{align*}
p(w,w') 
&=  p(w' \mid w) p(w)\\
&= \frac{X_{ww'}}{X_w} \frac{X_w}{\sum_{\hat{w}} X_{\hat{w}}}\\
&\propto X_{ww'} \,,
\end{align*} this gives us the Squared Norm (SN) objective (\ref{eq:objective-sq}).

%% file: methods.tex
\chapter{Methods}\label{chapter:methods}

\section{Training the word vectors}

Our training corpus $\mathcal{C}$ is an english Wikipedia corpus\footnote{\texttt{http://dumps.wikimedia.org/enwiki/latest/enwiki-latest-pages-articles.xml.bz2}} consisting of roughly 3.5 billion tokens, which was preprocessed before training so that multi-word named-entities (e.g., ``\texttt{princeton university}'') are treated as a single word (e.g., ``\texttt{princeton\_university}''); see Section \ref{section:preprocessing_corpus}.

We use a fixed context window of size $10$ to compute the co-occurrence data. Since it is too memory-expensive to store the co-occurrence count between all distinct words that appear in $\mathcal{C}$, and because the co-occurrence data for infrequent words are too noisy to generate good word vectors with, we fix a minimum threshold $m_0 \in \mathbb{N}$, and only consider the set $\mathcal{D}$ of vocabulary words that appear at least $m_0$ times in the training corpus.\footnote{Words which appear \emph{less} than $m_0=1000$ times are ignored in the sense that (1) co-occurrence counts for these words are not computed, and (2) these words are ignored when computing co-occurrence counts for other words. They do not affect or change the context window around any word (i.e., they are not deleted from the corpus).} In our experiments, we set $m_0 = 1000$, and the resulting vocabulary size is $|\mathcal{D}| = 60,265$. Words which appear fewer than 1000 times in the corpus are not included in the vocabulary $\mathcal{D}$, and hence, we do not learn vectors for these words. In Chapter \ref{chapter:fit_new_vec}, we provide a way of learning vectors for less frequent words with only a small amount of co-occurrence data.

To train the word vectors $V_{\mathcal{D}} := \{ v_w : w \in \mathcal{D}\}$, we optimize the Squared Norm (SN) objective (\ref{eq:objective-sq}) using Adagrad \cite{adagrad}. We use $d=300$ as the dimension of the word vectors $v_w \in \mathbb{R}^d$, which is also the dimension used in \cite{MSC+13, GloVe, arora} for their experiments on analogy query tasks. After training, we normalize the learned word vectors $v_w \in V_{\mathcal{D}}$ so that $\len{v_w} = 1$.

\section{Preprocessing the corpus}\label{section:preprocessing_corpus}

Word embeddings are inherently limited by their inability to represent multi-word, idiomatic phrases whose meanings are not simple compositions of the individual words. For instance, ``\texttt{kevi-n\_bacon}'' is the name of an individual, and so it is not a natural combination of the meanings of ``\texttt{kevin}'' and ``\texttt{bacon}''. Many facts about the world are concerned with multi-word entities, and hence, it is important to learn vectors for these entities.

To allow training of vectors for multi-word named-entities, we preprocessed the corpus $\mathcal{C}$ before training in the following manner: We used the named-entity recognition\footnote{Named-entity recognition (NER) is the task of labeling sequences of words in a text which are the names of things, such as the names of persons, organizations, and locations. NER is beyond the scope of this paper, and we refer the reader to \cite{NER}.} library in the Natural Language Toolkit (NLTK) \cite{NLTK} to identify strings $\xi$ which correspond to multi-word named-entities in the corpus, and replaced each space in $\xi$ with an underscore to make $\xi$ a single word. (For example, we replaced every instance of the string ``\texttt{princeton university}'' in the corpus with ``\texttt{princeton\_university}''.) After preprocessing, we train vectors for multi-word named-entities in the same way as with other words.

The preprocessing decreased the size of $\mathcal{D}$ (i.e., the number of vocabulary words that appear at least $m_0 = 1000$ times in the corpus) from $68,430$ to $60,265$, but \emph{increased} the number of words that appear at least $100$ times in the corpus from $296,376$ to $344,112$.

%% file: category_relations.tex
\chapter{Categories and relations}\label{chapter:category_relations}

In this chapter, we introduce the notion of \emph{categories} and \emph{relations}, which can be used to represent facts about the world in a knowledge base\footnote{Freebase \cite{freebase} is an example of a knowledge base whose data items are organized using categories and relations.}. In Figures \ref{fig:freebase_category} and \ref{fig:freebase_relation}, we list the categories and relations that are used for experiments in the subsequent chapters.

\section{Notation}

Let $\mathcal{D}$ be the set of vocabulary words that appear in a corpus $\mathcal{C}$. Words $w \in \mathcal{D}$ belong to certain \emph{categories}: for example, the word \texttt{princeton} belongs to the categories \texttt{university} and \texttt{municipality}, while the word \texttt{christianity} belongs to the category \texttt{religion}. Moreover, a pair of words sometimes satisfy a certain \emph{relation}: e.g., the word pair \texttt{(united\_states, dollar)} satisfies the relation \texttt{currency\_used}.

Given a category $c$, let $\mathcal{D}_c \subseteq \mathcal{D}$ be the set of words that belong to $c$, and let
\[ V_c := \{ v_w : w \in \mathcal{D}_c \}.\]
Similarly, given a relation $r$, let $\mathcal{D}_r \subseteq \mathcal{D} \times \mathcal{D}$ be the set of word pairs that satisfy the relation $r$, and let
\[ V_r := \{ v_a - v_b : (a,b) \in \mathcal{D}_r \}.\]
The vector subspace spanned by $V_c$ is called the \emph{subspace of category $c$}, and similarly, the vector subspace spanned by $V_r$ is called the \emph{subspace of relation $r$}. Moreover, we say a relation $r$ is \emph{well-defined} if there exists categories $c_A$ and $c_B$ such that, for all $(a,b) \in \mathcal{D}_r$, $a$ belongs to $c_A$ and $b$ belongs to $c_B$. \eat{Note that a well-defined relation can be thought of as the edges of a bipartite graph, where the bipartitions are $\mathcal{D}_{c_A}$ and $\mathcal{D}_{c_B}$ (see Figure \ref{fig:relation_types} for an example).}

For example, if $c = \texttt{country}$, then $\mathcal{D}_c$ contains words such as \texttt{germany}, \texttt{japan}, and \texttt{united\_states}. If $r = \texttt{capital\_city}$, then $\mathcal{D}_r$ contains pairs such as \texttt{(germany, berlin)}, \texttt{(japan, tokyo)}, and \texttt{(united\_states, washington\_dc)}. Also, the relation $r=\texttt{capital\_city}$ is well-defined, with $c_A = \texttt{country}$ and $c_B = \texttt{city}$. See Figure \ref{fig:relation_types}(a)-(h) for other examples of well-defined relations.

\section{Obtaining training examples of categories and relations}\label{section:categories_and_relations}

We obtained training examples of different categories and relations from various sources including Freebase \cite{freebase} and the GloVe demo code \cite{GloVe}. Figures \ref{fig:freebase_category} and \ref{fig:freebase_relation} list the category and relation files that are used for experiments in the subsequent chapters. Note that some words in the category and relation files, such as \texttt{maremma\_sheepdog} in the category \texttt{animal}, are not in the set $\mathcal{D}$ of vocabulary words because they appear fewer than $m_0=1000$ times in the corpus $\mathcal{C}$. Since we do not have learned vectors for these words, we remove them from the category and relation files in the experiments.

\input{freebase.tex}

\section{Experiments in the subsequent chapters}

In Chapter \ref{chapter:capture_rate}, we will show that the subspace of a category $c$, or the subspace of a well-defined relation $r$, can be approximated well by a relatively low-dimensional subspace, whose basis vectors can be computed using SVD. In Chapter \ref{chapter:learn_new_facts}, we use this basis to discover new words belonging to a category, or new word pairs satisfying a well-defined relation. In Chapter \ref{chapter:fit_new_vec}, we also use this basis to learn vectors for words with substantially less co-occurrence data.

Chapter \ref{chapter:wordnet_filter} explores the idea of using external knowledge sources such as Wordnet \cite{wn} to improve accuracy on analogy queries, and uses the relation files in Figure \ref{fig:freebase_relation} as a benchmark.

%% file: freebase.tex
%\chapter{Categories and relations}

\RecustomVerbatimCommand{\VerbatimInput}{VerbatimInput}%
{fontsize=\footnotesize,
firstline=1,
lastline=5,
frame=single,
boxwidth=5cm,
labelposition=topline
}
\begin{figure}
\input{freebase_category.tex}
\caption[Category files]{For each category file, we list the first 6 words and the total number of words it contains. \label{fig:freebase_category}}
\end{figure}

%\begin{figure}
%\ContinuedFloat
%\end{figure}

\RecustomVerbatimCommand{\VerbatimInput}{VerbatimInput}%
{fontsize=\footnotesize,
firstline=1,
lastline=9,
frame=single,
boxwidth=5cm,
labelposition=topline
}

\begin{figure}
\input{freebase_relation-facts.tex}
\caption[Facts-based relation files]{A list of facts-based relation files. For each relation file, we list the first 9 word pairs and the total number of word pairs it contains. Note that (a) is a smaller subset of (h), and (e) is a smaller subset of (g), the former containing only the more ``well-known'' pairs.}
\end{figure}

\begin{figure}
\ContinuedFloat
\input{freebase_relation-gram.tex}
\caption[Grammar-based relation files]{A list of grammar-based relation files. For each relation file, we list the first 9 word pairs and the total number of word pairs it contains. }
\end{figure}

\begin{figure}
\ContinuedFloat
\input{freebase_relation-semantics.tex}
\caption[Semantics-based relation files]{A list of semantics-based relation files. For each relation file, we list the first 9 word pairs and the total number of word pairs it contains. \label{fig:freebase_relation}}
\end{figure}

%% file: freebase_category.tex
\begin{subfigure}{0.33\linewidth}
\begin{minipage}{\linewidth}
\VerbatimInput{animal.txt}
\end{minipage}
\caption{\footnotesize\texttt{animal} (927 words)}
\end{subfigure}\hspace{1ex}%%%%%%%%%
\begin{subfigure}{0.33\linewidth}
\begin{minipage}{\linewidth}
\VerbatimInput{asian_city}
\end{minipage}
\caption{\footnotesize\texttt{asian\_city} (15 words)}
\end{subfigure}\hspace{1ex}%%%%%%%%%
\begin{subfigure}{0.33\linewidth}
\begin{minipage}{\linewidth}
\VerbatimInput{basketball_player.txt}
\end{minipage}
\caption{\footnotesize\texttt{basketball\_player} (45 words)}
\end{subfigure}%\hspace{1ex}%%%%%%%%%

\vspace{2ex}
\begin{subfigure}{0.33\linewidth}
\begin{minipage}{\linewidth}
\VerbatimInput{cheese.txt}
\end{minipage}
\caption{\footnotesize\texttt{cheese} (518 words)}
\end{subfigure}\hspace{1ex}%%%%%%%%%
\begin{subfigure}{0.33\linewidth}
\begin{minipage}{\linewidth}
\VerbatimInput{capital_country.txt-2}
\end{minipage}
\caption{\footnotesize\texttt{city} (560 words)}
\end{subfigure}\hspace{1ex}%%%%%%%%%
\begin{subfigure}{0.33\linewidth}
\begin{minipage}{\linewidth}
\VerbatimInput{classical_composer.txt}
\end{minipage}
\caption{\footnotesize\texttt{classical\_composer} (7 words)}
\end{subfigure}%\hspace{1ex}%%%%%%%%%

\vspace{2ex}
\begin{subfigure}{0.33\linewidth}
\begin{minipage}{\linewidth}
\VerbatimInput{cocktail.txt}
\end{minipage}
\caption{\footnotesize\texttt{cocktail} (386 words)}
\end{subfigure}\hspace{1ex}%%%%%%%%%
\begin{subfigure}{0.33\linewidth}
\begin{minipage}{\linewidth}
\VerbatimInput{capital_country.txt-1}
\end{minipage}
\caption{\footnotesize\texttt{country} (643 words)}
\end{subfigure}\hspace{1ex}%%%%%%%%%
\begin{subfigure}{0.33\linewidth}
\begin{minipage}{\linewidth}
\VerbatimInput{currency_used-2}
\end{minipage}
\caption{\footnotesize\texttt{currency} (112 words)}
\end{subfigure}%\hspace{1ex}%%%%%%%%%

\vspace{2ex}
\begin{subfigure}{0.33\linewidth}
\begin{minipage}{\linewidth}
\VerbatimInput{holiday.txt}
\end{minipage}
\caption{\footnotesize\texttt{holiday} (956 words)}
\end{subfigure}\hspace{1ex}%%%%%%%%%
\begin{subfigure}{0.33\linewidth}
\begin{minipage}{\linewidth}
\VerbatimInput{languages.txt-2}
\end{minipage}
\caption{\footnotesize\texttt{languages} (534 words)}
\end{subfigure}\hspace{1ex}%%%%%%%%%
\begin{subfigure}{0.33\linewidth}
\begin{minipage}{\linewidth}
\VerbatimInput{month}
\end{minipage}
\caption{\footnotesize\texttt{month} (12 words)}
\end{subfigure}\hspace{1ex}%%%%%%%%%

\vspace{2ex}
\begin{subfigure}{0.33\linewidth}
\begin{minipage}{\linewidth}
\VerbatimInput{musical_instrument.txt}
\end{minipage}
\caption{\footnotesize\footnotesize\texttt{musical\_instrument} (994 words)}
\end{subfigure}\hspace{1ex}%%%%%%%%%
\begin{subfigure}{0.33\linewidth}
\begin{minipage}{\linewidth}
\VerbatimInput{music_genre.txt}
\end{minipage}
\caption{\footnotesize\texttt{music\_genre} (997 words)}
\end{subfigure}\hspace{1ex}%%%%%%%%%
\begin{subfigure}{0.33\linewidth}
\begin{minipage}{\linewidth}
\VerbatimInput{organism_classification.txt}
\end{minipage}
\caption{\footnotesize\texttt{organism\_group} (999 words)}
\end{subfigure}%\hspace{1ex}%%%%%%%%%

\vspace{2ex}
\begin{subfigure}{0.33\linewidth}
\begin{minipage}{\linewidth}
\VerbatimInput{politician.txt}
\end{minipage}
\caption{\footnotesize\texttt{politician} (1786 words)}
\end{subfigure}\hspace{1ex}%%%%%%%%%
\begin{subfigure}{0.33\linewidth}
\begin{minipage}{\linewidth}
\VerbatimInput{religion.txt}
\end{minipage}
\caption{\footnotesize\texttt{religion} (593 words)}
\end{subfigure}\hspace{1ex}%%%%%%%%%
\begin{subfigure}{0.33\linewidth}
\begin{minipage}{\linewidth}
\VerbatimInput{sport.txt}
\end{minipage}
\caption{\footnotesize\texttt{sport} (686 words)}
\end{subfigure}%\hspace{1ex}%%%%%%%%%

\vspace{2ex}
\begin{subfigure}{0.33\linewidth}
\begin{minipage}{\linewidth}
\VerbatimInput{tourist_attraction.txt}
\end{minipage}
\caption{\footnotesize\texttt{tourist\_attraction} (9905 words)}
\end{subfigure}\hspace{1ex}%%%%%%%%%
\begin{subfigure}{0.33\linewidth}
\begin{minipage}{\linewidth}
\VerbatimInput{university}
\end{minipage}
\caption{\footnotesize\texttt{university} (15 words)}
\end{subfigure}\hspace{1ex}

%% file: freebase_relation-facts.tex
%\vspace{2ex}
\begin{subfigure}{0.5\linewidth}
\begin{minipage}{\linewidth}
\VerbatimInput{currency}
\end{minipage}
\caption{\footnotesize\texttt{country-currency} (30 pairs)}
\end{subfigure}\hspace{1ex}%%%%%%%%%
\begin{subfigure}{0.5\linewidth}
\begin{minipage}{\linewidth}
\VerbatimInput{holiday-month}
\end{minipage}
\caption{\footnotesize\texttt{holiday-month} (4 pairs)}
\end{subfigure}%\hspace{1ex}%%%%%%%%%

\vspace{2ex}
\begin{subfigure}{0.5\linewidth}
\begin{minipage}{\linewidth}
\VerbatimInput{holiday-religion}
\end{minipage}
\caption{\footnotesize\texttt{holiday-religion} (58 pairs)}
\end{subfigure}\hspace{1ex}%%%%%%%%%
\begin{subfigure}{0.5\linewidth}
\begin{minipage}{\linewidth}
\VerbatimInput{language}
\end{minipage}
\caption{\footnotesize\texttt{country-language} (38 pairs)}
\end{subfigure}%\hspace{1ex}%%%%%%%%%

\vspace{2ex}
\begin{subfigure}{0.5\linewidth}
\begin{minipage}{\linewidth}
\VerbatimInput{capital-all}
\end{minipage}
\caption{\footnotesize\texttt{country-capital2} (63 pairs)}
\end{subfigure}\hspace{1ex}%%%%%%%%%
\begin{subfigure}{0.5\linewidth}
\begin{minipage}{\linewidth}
\VerbatimInput{city-in-state}
\end{minipage}
\caption{\footnotesize\texttt{city-in-state} (87 pairs)}
\end{subfigure}%\hspace{1ex}%%%%%%%%%

\vspace{2ex}
\begin{subfigure}{0.5\linewidth}
\begin{minipage}{\linewidth}
\VerbatimInput{capital_country.txt}
\end{minipage}
\caption{\footnotesize\texttt{capital\_country} (869 pairs)}
\end{subfigure}\hspace{1ex}%%%%%%%%%
\begin{subfigure}{0.5\linewidth}
\begin{minipage}{\linewidth}
\VerbatimInput{currency_used}
\end{minipage}
\caption{\footnotesize\texttt{currency\_used} (381 pairs)}
\end{subfigure}%\hspace{1ex}%%%%%%%%%

%% file: freebase_relation-gram.tex
%\vspace{2ex}
\begin{subfigure}{0.5\linewidth}
\begin{minipage}{\linewidth}
\VerbatimInput{gram1-adjective-to-adverb}
\end{minipage}
\caption{\footnotesize\texttt{gram1-adj-adv} (32 pairs)}
\end{subfigure}\hspace{1ex}%%%%%%%%%
\begin{subfigure}{0.5\linewidth}
\begin{minipage}{\linewidth}
\VerbatimInput{gram2-opposite}
\end{minipage}
\caption{\footnotesize\texttt{gram2-opposite} (29 pairs)}
\end{subfigure}%\hspace{1ex}%%%%%%%%%

\vspace{2ex}
\begin{subfigure}{0.5\linewidth}
\begin{minipage}{\linewidth}
\VerbatimInput{gram3-comparative}
\end{minipage}
\caption{\footnotesize\texttt{gram3-comparative} (37 pairs)}
\end{subfigure}\hspace{1ex}%%%%%%%%%
\begin{subfigure}{0.5\linewidth}
\begin{minipage}{\linewidth}
\VerbatimInput{gram4-superlative}
\end{minipage}
\caption{\footnotesize\texttt{gram4-superlative} (34 pairs)}
\end{subfigure}%\hspace{1ex}%%%%%%%%%

\vspace{2ex}
\begin{subfigure}{0.5\linewidth}
\begin{minipage}{\linewidth}
\VerbatimInput{gram5-present-participle}
\end{minipage}
\caption{\footnotesize\texttt{gram5-present-participle} (33 pairs)}
\end{subfigure}\hspace{1ex}%%%%%%%%%
\begin{subfigure}{0.5\linewidth}
\begin{minipage}{\linewidth}
\VerbatimInput{gram6-nationality-adjective}
\end{minipage}
\caption{\footnotesize\texttt{gram6-nationality-adj} (40 pairs)}
\end{subfigure}%\hspace{1ex}%%%%%%%%%

\vspace{2ex}
\begin{subfigure}{0.5\linewidth}
\begin{minipage}{\linewidth}
\VerbatimInput{gram7-past-tense}
\end{minipage}
\caption{\footnotesize\texttt{gram7-past-tense} (40 pairs)}
\end{subfigure}\hspace{1ex}%%%%%%%%%
\begin{subfigure}{0.5\linewidth}
\begin{minipage}{\linewidth}
\VerbatimInput{gram8-plural}
\end{minipage}
\caption{\footnotesize\texttt{gram8-plural} (37 pairs)}
\end{subfigure}%\hspace{1ex}%%%%%%%%%

\vspace{2ex}
\begin{subfigure}{0.5\linewidth}
\begin{minipage}{\linewidth}
\VerbatimInput{gram9-plural-verbs}
\end{minipage}
\caption{\footnotesize\texttt{gram9-plural-verbs} (30 pairs)}
\end{subfigure}\hspace{1ex}%%%%%%%%%

%% file: freebase_relation-semantics.tex
%\vspace{2ex}
\begin{subfigure}{0.5\linewidth}
\begin{minipage}{\linewidth}
\VerbatimInput{associated-number}
\end{minipage}
\caption{\footnotesize\texttt{associated-number} (18 pairs)}
\end{subfigure}\hspace{1ex}%%%%%%%%%
\begin{subfigure}{0.5\linewidth}
\begin{minipage}{\linewidth}
\VerbatimInput{associated-position}
\end{minipage}
\caption{\footnotesize\texttt{associated-position} (10 pairs)}
\end{subfigure}%\hspace{1ex}%%%%%%%%%

\vspace{2ex}
\begin{subfigure}{0.5\linewidth}
\begin{minipage}{\linewidth}
\VerbatimInput{common-very}
\end{minipage}
\caption{\footnotesize\texttt{common-very} (10 pairs)}
\end{subfigure}\hspace{1ex}%%%%%%%%%
\begin{subfigure}{0.5\linewidth}
\begin{minipage}{\linewidth}
\VerbatimInput{graded-antonym}
\end{minipage}
\caption{\footnotesize\texttt{graded-antonym} (37 pairs)}
\end{subfigure}%\hspace{1ex}%%%%%%%%%

\vspace{2ex}
\begin{subfigure}{0.5\linewidth}
\begin{minipage}{\linewidth}
\VerbatimInput{has-characteristic}
\end{minipage}
\caption{\footnotesize\texttt{has-characteristic} (4 pairs)}
\end{subfigure}\hspace{1ex}%%%%%%%%%
\begin{subfigure}{0.5\linewidth}
\begin{minipage}{\linewidth}
\VerbatimInput{has-function}
\end{minipage}
\caption{\footnotesize\texttt{has-function} (19 pairs)}
\end{subfigure}%\hspace{1ex}%%%%%%%%%

\vspace{2ex}
\begin{subfigure}{0.5\linewidth}
\begin{minipage}{\linewidth}
\VerbatimInput{has-skin}
\end{minipage}
\caption{\footnotesize\texttt{has-skin} (8 pairs)}
\end{subfigure}\hspace{1ex}%%%%%%%%%
\begin{subfigure}{0.5\linewidth}
\begin{minipage}{\linewidth}
\VerbatimInput{noun-baby}
\end{minipage}
\caption{\footnotesize\texttt{noun-baby} (12 pairs)}
\end{subfigure}%\hspace{1ex}%%%%%%%%%

\vspace{2ex}
\begin{subfigure}{0.5\linewidth}
\begin{minipage}{\linewidth}
\VerbatimInput{senses}
\end{minipage}
\caption{\footnotesize\texttt{senses} (5 pairs)}
\end{subfigure}\hspace{1ex}%%%%%%%%%

%% file: capture_rate.tex
\chapter{Low-rank subspaces of categories and relations}\label{chapter:capture_rate}
\chaptermark{Low-rank subspaces}

We demonstrate that the subspace of a category $c$, or the subspace of a well-defined relation $r$, can be approximated well by a relatively low-dimensional subspace in the projected space. More specifically, we use singular value decomposition (SVD) to approximate a low-rank basis $U_k = \{ u_1, \ldots, u_k \}$ for the subspace of category $c$ or relation $r$ (see Algorithm \ref{section:svd_basis}). Section \ref{section:experiment-capture_rate} outlines an experiment where we use cross-validation to show that $U_k$ captures most of the vectors in $V_c = \{ v_w : w \in \mathcal{D}_c \}$ or $V_r = \{ v_a - v_b : (a,b) \in \mathcal{D}_r \}$ (see Figure \ref{fig:svd_capture_rate}). Moreover, we show that the first basis vector $u_1$ captures the most information about a category $c$ or a relation $r$ (see Figures \ref{fig:svd_capture_direction} and \ref{fig:svd_capture_direction-relations}).

%%%%%%%%%%%%%%%%%%%%%%%%%%%%%%%%%%%%%%%%%%%%%%%%%%%%%%%%%%%%%%%%%%%%%%%%%%%%%%%%
\renewcommand{\figurename}{Algorithm}
\begin{figure}
\begin{framed}
\noindent {\bf function $\textsf{GET\_BASIS}( V, k )$}: Returns a rank-$k$ basis for the subspace spanned by $V$.\\\\
\noindent Inputs:
\begin{itemize}
\item $V = \{ v_1, v_2, \ldots, v_n \}$, a set of vectors in $\mathbb{R}^d$. Let $|V| := n$ denote the number of vectors in $V$.
\item $k \in \mathbb{N}$, the rank of the basis (where $k \ll d$)
\end{itemize}

\begin{enumerate}[Step 1.]
\item Let $X$ be a $d \times |V|$ matrix whose column vectors are $v \in V$. Using SVD, factorize $X$ as
\[
X = U \Sigma W^T,
\]
where $U \in \mathbb{R}^{d \times d}$ and $W \in \mathbb{R}^{|V| \times |V|}$ are orthogonal matrices, and $\Sigma \in \mathbb{R}^{d \times |V|}$ is the diagonal matrix of the singular values of $X$ in descending order.

\item Let $U_k$ be the $d \times k$ submatrix obtained by taking the first $k$ columns $u_1, \ldots, u_k$ of $U$ (which correspond to the $k$ largest singular values of $X$). Since $U$ is orthogonal, the columns of $U_k$ form a rank-$k$ orthonormal basis.
\item Scale $u_1$ by $\pm 1$ so that $\sum_{v \in V} v \cdot u_1 > 0$.

\item Return $U_k$.
\end{enumerate}

\noindent {\bf end function}
\end{framed}
\caption{$\textsf{GET\_BASIS}(V,k)$ returns a rank-$k$ basis, $U_k \in \mathbb{R}^{d \times k}$, for the subspace spanned by a set of vectors $V$.}
\label{section:svd_basis}
\end{figure}
\renewcommand{\figurename}{Figure}
%%%%%%%%%%%%%%%%%%%%%%%%%%%%%%%%%%%%%%%%%%%%%%%%%%%%%%%%%%%%%%%%%%%%%%%%%%%%%%%%

\section{Computing a low-rank basis using SVD}

The function \textsf{GET\_BASIS} (Algorithm \ref{section:svd_basis}) uses SVD to compute a rank-$k$ basis for the subspace spanned by a set of vectors $V$. So $\textsf{GET\_BASIS}(V_c,\, k)$ and $\textsf{GET\_BASIS}(V_r,\, k)$ return a rank-$k$ basis for the subspace of category $c$ and for the subspace of relation $r$, respectively.

\begin{center}\begin{figure}
\includegraphics[width=\textwidth]{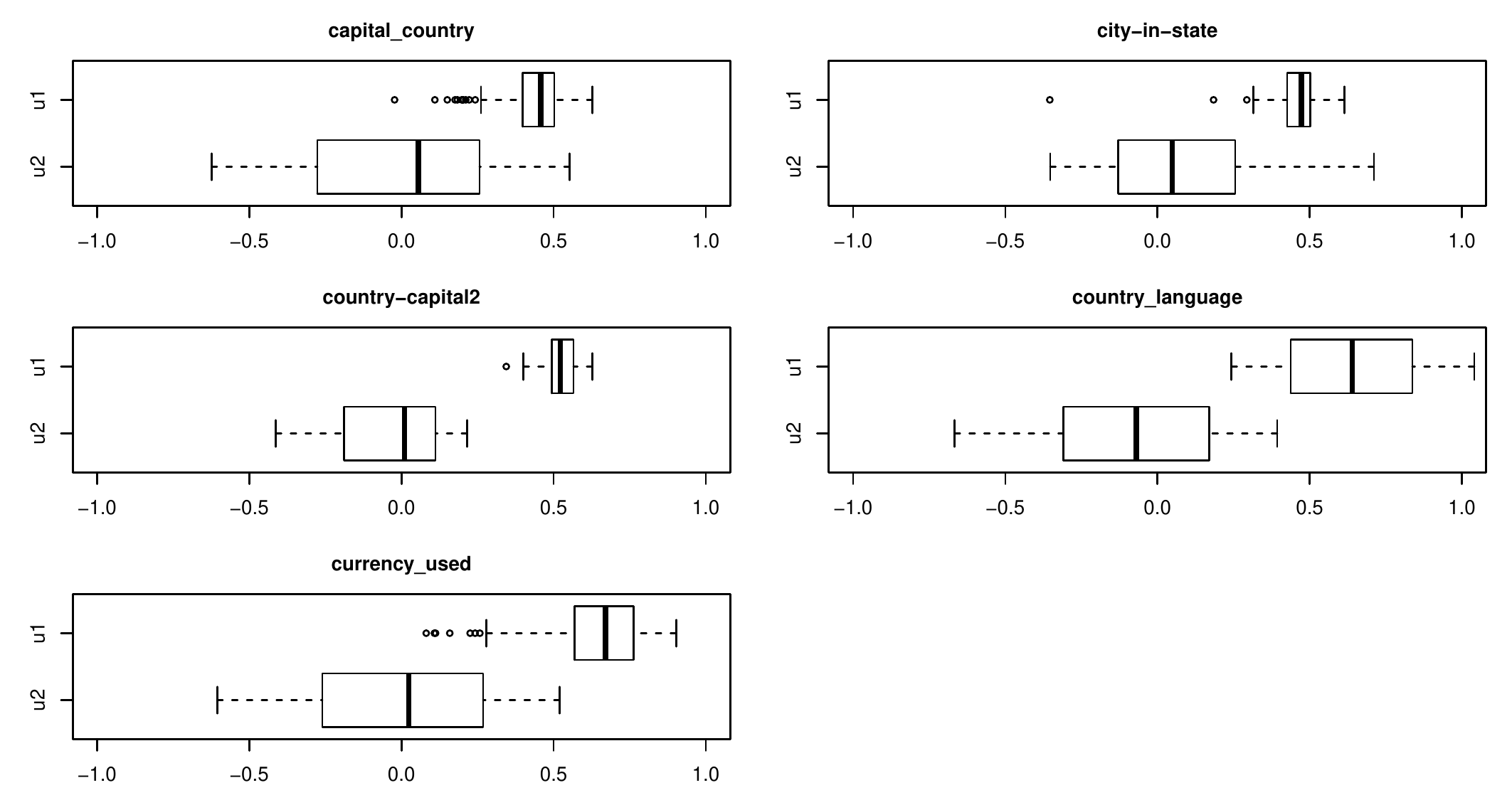}
\caption{For various relations $r$ from Figure \ref{fig:freebase_relation}, we randomly split $V_r$ into a training set $V_1$ and a testing set $V_2$, then compute a rank-$k$ basis $U_k = \{ u_1, u_2, \ldots, u_k \}$ for the subspace spanned by $V_1$. Here, we provide a boxplot of the distribution of $\{ v \cdot u_1 \}_{v \in V_2}$ (denoted by \texttt{u1}) and the distribution of $\{ v \cdot u_2 \}_{v \in V_2}$ (denoted by \texttt{u2}). Note that for eachrelation $r$, we have $v \cdot u_1 > 0$ for all $v \in V_2$ (except for one outlier in $r = \texttt{capital\_country}$, and one outlier in $r=\texttt{city-in-state}$), whereas $\{ v \cdot u_2 \}_{v \in V_2}$ is more evenly distributed around 0. The distribution of $\{ v \cdot u_i \}_{v \in V_2}$ for $i \geq 2$ is similar to the distribution for $i=2$, so we omit them here.
\label{fig:svd_capture_direction-relations}}
\end{figure}\end{center}

In the following experiment, we use cross-validation to assess how much the low-rank basis $U_k = \{u_1, \ldots, u_k\}$ returned by \textsf{GET\_BASIS} captures the subspace of a category $c$ or a relation $r$. We demonstrate that the first basis vector $u_1$ (corresponding to the largest singular value $\sigma_1$) captures the most information. In particular, we show that the first basis vector $u_1$ captures significantly more mass of the subspace than any other basis vector (see Figure \ref{fig:svd_capture_rate}), and that the only $i \in \{ 1, \ldots, k \}$ satisfying
\[ v \cdot u_i > 0 \quad \forall v \in V_c \quad\textrm{(or $\forall v \in V_r$)}\]
is $i=1$ (see Figures \ref{fig:svd_capture_direction} and \ref{fig:svd_capture_direction-relations}).

\subsection{Experiment}\label{section:experiment-capture_rate}

Let $c$ be a category where we have a set $S_c \subseteq \mathcal{D}_c$ of words that belong to the category $c$, with size $|S_c| > 50$. For each rank $k \in \{ 1, 2, \ldots, 25 \}$, we perform the following experiment\footnote{The same experiment is done for a relation $r$, by replacing $V_c$ with $V_r$, and $S_c \subseteq \mathcal{D}_c$ with $S_r \subseteq \mathcal{D}_r$.} over $T=50$ repeated trials:

\begin{enumerate}[Step 1.]
\item Randomly partition $S_c$ into a training set $S_1$ and a testing set $S_2$, where the training set size is $|S_1| = 0.7 |S_c|$. For each $i \in \{1,2\}$, let $V_i := \{ v_w : w \in S_i \}$.

\item Compute a rank-$k$ basis $U_k$ for the subspace spanned by $V_1$, using Algorithm \ref{section:svd_basis}:
\[ U_k \leftarrow \textrm{GET\_BASIS}(V_1, k).\]

\item To measure how much a vector $v \in \mathbb{R}^d$ is captured by the span of $U_k \in \mathbb{R}^{d \times k}$, define
\[ \phi(U_k,v) := \frac{\len{U_k^T v}}{\len{v}}. \]
Now, test how much $U_k$ captures the vectors $V_2$ of the testing set, by computing the \emph{capture rate}
\begin{equation}
\phi(U_k, V_2) := {1 \over |V_2|} \sum_{v \in V_2} \phi(U_k, v).
\label{eq:capture_rate}
\end{equation}
If $\phi(U_k, V_2)$ is large, i.e., the vectors in $V_2$ have a large projection onto $U_k$, then it would indicate that $U_k$ is a good low-rank approximation for the subspace of category $c$.

\item Look at the distribution of the values in $\{ u_i \cdot v \}_{v \in V_2}$ for each $i \in \{1, \ldots, k\}$.
\end{enumerate}

\section{Results}

\eat{For each rank, we repeated the experiment in Section \ref{section:experiment-capture_rate} for $T=50$ trials.}

\begin{center}\begin{figure}
\includegraphics[width=\textwidth]{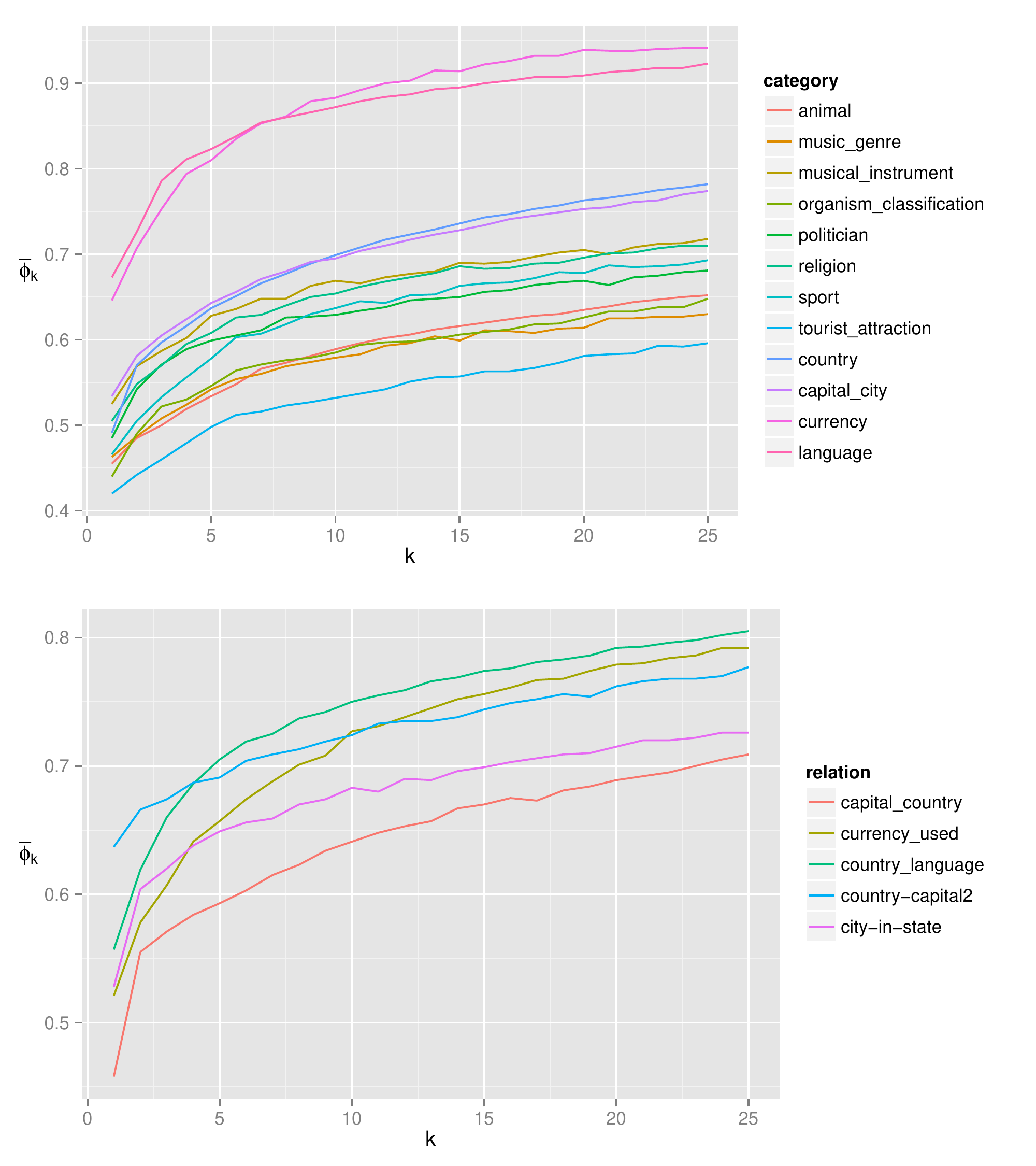}
\caption{Results from the experiment in Section \ref{section:experiment-capture_rate}, where we use cross-validation over $T=50$ repeated trials to assess how much the low-rank basis $U_k$ returned by \textsf{GET\_BASIS} (Algorithm \ref{section:svd_basis}) captures the subspace of a category $c$ (or a relation $r$). In each trial $t \in \{1, \ldots, T\}$, we randomly split $V_c$ (or $V_r$) into a training set $V_1^{(t)}$ and a testing set $V_2^{(t)}$, then compute a rank-$k$ basis $U_k^{(t)}$ for the subspace spanned by $V_1^{(t)}$. For each rank $k \in \{1, 2, \ldots 25\}$, we plot the average capture rate $\bar{\phi}_k := {1 \over T} \sum_{t=1}^T \phi(U_k^{(t)}, V_2^{(t)})$ (\ref{eq:capture_rate}), where a higher capture rate means that the vectors in $V_2$ have a large projection onto $U_k$. When rank is $k=1$, $\bar{\phi}_k$ is already between $0.420$ and $0.673$. For ranks $k \geq 2$ on the other hand, the increase from $\bar{\phi}_{k-1}$ to $\bar{\phi}_k$ is relatively small. This shows that the first basis vector $u_1$ is the ``defining'' vector that encodes the most information about a category $c$ (or a relation $r$).\label{fig:svd_capture_rate}}
\end{figure}\end{center}

For each rank $k \in \{1, 2, \ldots 25\}$, we plot the average capture rate $\bar{\phi}_k := {1 \over T} \sum_{t=1}^T \phi(U_k^{(t)}, V_2^{(t)})$ over $T=50$ repeated trials\footnote{For each trial $t \in \{1, \ldots, T\}$, $\phi(U_k^{(t)}, V_2^{(t)})$ is the capture rate attained in trial $t$, where $V_2^{(t)} = \{ v_w : w \in S_2^{(t)} \}$ is the set of vectors for words in the testing set $S_2^{(t)}$ (which is randomly generated in Step 1), and $U_k^{(t)}$ is the rank-$k$ basis computed in Step 2.} in Figure \ref{fig:svd_capture_rate}. Notice that when $k=1$, $\bar{\phi}_k$ is already between $0.420$ and $0.673$. For $k \geq 2$ on the other hand, the increase from $\bar{\phi}_{k-1}$ to $\bar{\phi}_k$ is relatively small.

\begin{center}\begin{figure}
\includegraphics[width=\textwidth]{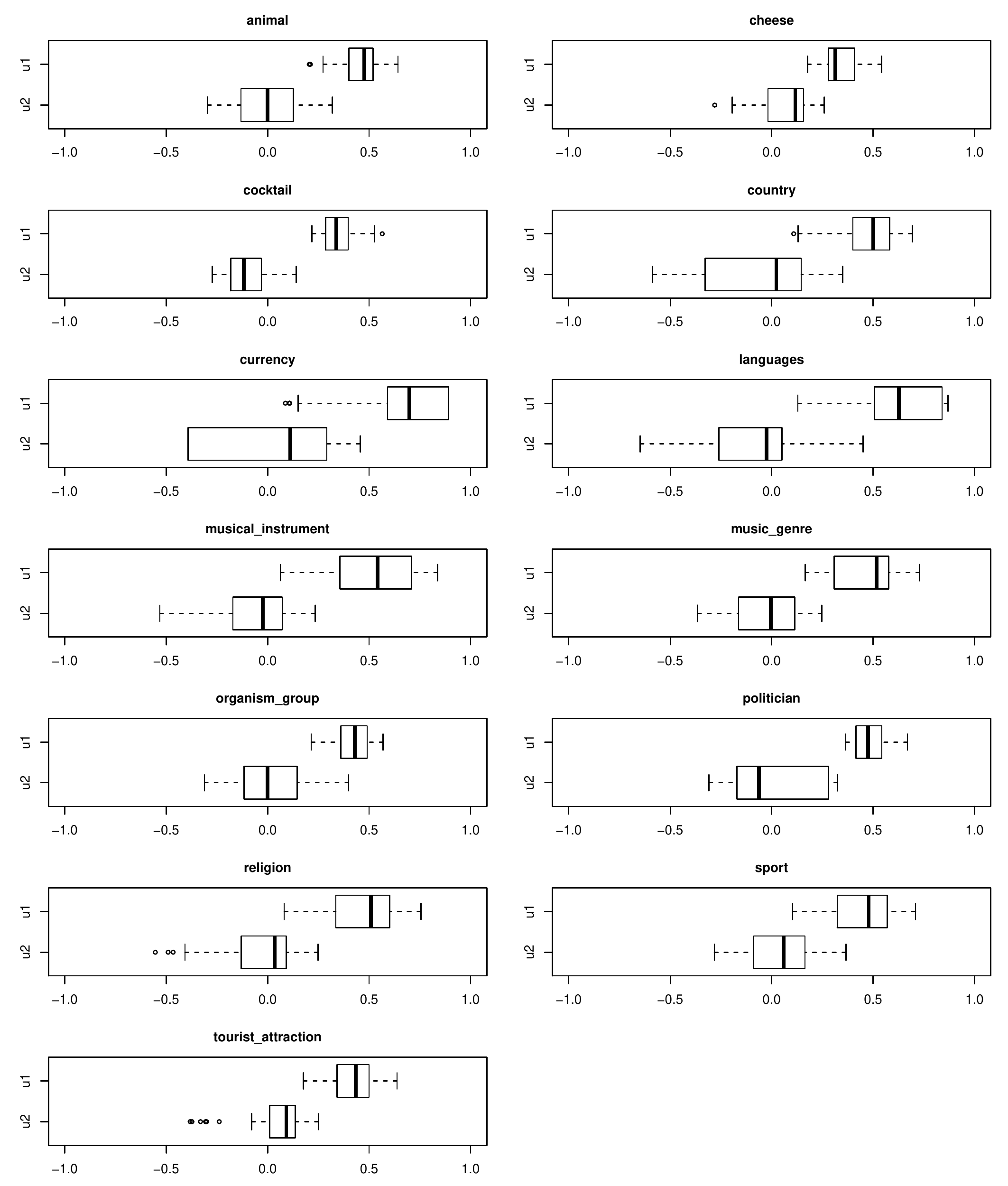}
\caption{For various categories $c$ from Figure \ref{fig:freebase_category}, we randomly split $V_c$ into a training set $V_1$ and a testing set $V_2$, then compute a rank-$k$ basis $U_k = \{ u_1, u_2, \ldots, u_k \}$ for the subspace spanned by $V_1$. Here, we provide a boxplot of the distribution of $\{ v \cdot u_1 \}_{v \in V_2}$ (denoted by \texttt{u1}) and the distribution of $\{ v \cdot u_2 \}_{v \in V_2}$ (denoted by \texttt{u2}). Note that for each category $c$, we have $v \cdot u_1 > 0$ for all $v \in V_2$, whereas $\{ v \cdot u_2 \}_{v \in V_2}$ is more evenly distributed around 0. The distribution of $\{ v \cdot u_i \}_{v \in V_2}$ for $i \geq 2$ is similar to the distribution for $i=2$, so we omit them here.
\label{fig:svd_capture_direction}}
\end{figure}\end{center}

We found that the values $\{ v \cdot u_1\}_{v \in V_2}$ all have the same sign\footnote{Recall that, in Step 3 of Algorithm \ref{section:svd_basis}, we scale $u_1$ by $\pm 1$ so that $\sum_{v \in V_1} v \cdot u_1 > 0$.}, whereas for $i \geq 2$, the values $\{ v \cdot u_i\}_{v \in V_2}$ are more evenly distributed around 0. Figure \ref{fig:svd_capture_direction} shows the distribution $\{ v \cdot u_i\}_{v \in V_2}$ for $i=1$ and $i=2$; the distribution $\{ v \cdot u_i\}_{v \in V_2}$ for $i \geq 2$ is similar to the distribution for $i=2$.

\section{Conclusion}

Our results show that the subspace of many categories and relations are low-dimensional. Moreover, we demonstrated that the first basis $u_1$ is the ``defining'' vector that  encodes the most \emph{general} information about a category $c$ (or a relation $r$): Letting $v = v_w$ for any $w \in \mathcal{D}_c$ (or $v=v_a-v_b$ for any $(a,b) \in \mathcal{D}_r$), the first coordinate $v \cdot u_1$ has the largest magnitude, and the sign of $v \cdot u_1$ is always positive. All other subsequent basis vectors $u_i$ for $i \geq 2$ encode more ``specific'' information pertaining to individual words in $\mathcal{D}_c$ (or word pairs in $\mathcal{D}_r$).

It remains to be discovered what specific features are captured by these basis vectors for various categories and relations. For example, if $U_k$ is a basis for the subspace of category $c=\texttt{country}$, then perhaps having a positive second coordinate $v_w \cdot u_2$ in the basis indicates that $w$ is a \emph{developed} country, and having a negative fourth coordinate $v_w \cdot u_4$ indicates that country $w$ is located in Europe. We leave this to future work.

%% file: learn_new_facts.tex
\chapter{Extending a knowledge base}\label{chapter:learn_new_facts}

Let $\KB$ denote the current knowledge base, which consists of facts about the world that the machine currently knows. We assume that $\KB$ is incomplete, and that there are new facts to be learned. In other words, there exists categories $c$ such that $\KB$ only knows a subset $S_c \subset \mathcal{D}_c$ of words that belong to $c$, and similarly, there exists relations $r$ such that $\KB$ only knows a subset $S_r \subset \mathcal{D}_r$ of word pairs that satisfy $r$. Our goal is to discover new facts outside $\KB$, such as words in $\mathcal{D} \backslash S_c$ that also belong to the category $c$, or word pairs in $(\mathcal{D} \times \mathcal{D}) \backslash S_r$ that also satisfy the relation $r$.

In Section \ref{section:learning_new_words_in_category}, we provide an algorithm called \textsf{EXTEND\_CATEGORY} for discovering words in $\mathcal{D}\backslash S_c$ that also belong to $c$ (see Algorithm \ref{section:discover_new_entities-algo}). Table \ref{table:learn_new_entities} lists the top 5 words returned by \textsf{EXTEND\_CATEGORY} for various categories, and shows that the algorithm performs very well.  Then in Section \ref{section:learning_new_pairs_in_relation}, we present an algorithm called \textsf{EXTEND\_RELATION} for discovering new word pairs in $(\mathcal{D} \times \mathcal{D})\backslash S_r$ that also satisfy $r$ (see Algorithm \ref{section:learn_new_relations-algo}). Its accuracy for returning correct word pairs is provided in Figure \ref{fig:discover_new_facts-performance}.

We demonstrate that the low-dimensional subspace of categories and relations can be used to discover new facts with fairly low false-positive rates. The performance of \textsf{EXTEND\_RELATION} (Algorithm \ref{section:learn_new_relations-algo}) is especially surprising, given the simplicity of the algorithm and the fact that it returns plausible word pairs out of all $|\mathcal{D}|(|\mathcal{D}| - 1) \approx 3.6\textnormal{e}\!+\!09$ possible word pairs in $\mathcal{D} \times \mathcal{D}$.

\section{Motivation}

In Socher et. al's paper \cite{ntn}, a neural tensor network (NTN) model is used to learn semantic word vectors that can capture relationships between two words. More specifically, their NTN model computes a score of how plausible a word pair $(a,b)$ satisfies a certain relationship $r$ by the following function:
\begin{equation}g(v_a, r, v_b) = b \cdot f\paren*{ v_a^T W_r^{[1:m]}v_b + \theta_r \begin{bmatrix}v_a\\v_b\end{bmatrix} + c_r }\label{eq:ntn}\end{equation}
where $v_a, v_b \in \mathbb{R}^d$ are the vector representations of the words $a,b$ respectively, $f=\tanh$ is a sigmoid function, and $W_r^{[1:m]} \in \mathbb{R}^{d \times d \times m}$ is a tensor. The remaining parameters for relation $r$ are the standard form of a neural network: $\theta_r \in \mathbb{R}^{m \times 2d}$ and $b, c_r \in \mathbb{R}^m$.

With this model, however, we see a potential danger of overfitting the data because the number of parameters in the term $v_a^T W_r^{[1:m]}v_b$ in (\ref{eq:ntn}) is \emph{quadratic} in $d$. Hence, our motivation is to employ the linear structure of the word embeddings to fit a more general model. The low-dimensional subspace of categories, as demonstrated in Chapter \ref{chapter:capture_rate}, gives rise to a simple algorithm that allows one to discover new facts that are not in $\KB$.

\section{Learning new words in a category}\label{section:learning_new_words_in_category}

Let $c$ be a category such that $\KB$ only knows a subset $S_c \subset \mathcal{D}_c$ of words that belong to $c$. We provide a method called \textsf{EXTEND\_CATEGORY} (Algorithm \ref{section:discover_new_entities-algo}) for discovering new words in $\mathcal{D}\backslash S_c$ that also belong to $c$.

%%%%%%%%%%%%%%%%%%%%%%%%%%%%%%%%%%%%%%%%%%%%%%%%%%%%%%%%%%%%%%%%%%%%%%%%%%%%%%%%
\renewcommand{\figurename}{Algorithm}
\begin{figure}
\begin{framed}
\noindent {\bf function \textsf{EXTEND\_CATEGORY}($S_c$, $k$, $\delta$)}: Returns a set of words in $\mathcal{D} \backslash S_c$.\\\\
\noindent Inputs:
\begin{itemize}
\item A subset $S_c \subset \mathcal{D}_c$ of words that belong to some category $c$
\item $k \in \mathbb{N}$, the rank of the basis for the subspace of category $c$
\item $\delta \in [0, 1]$, a threshold value
\eat{\item $\{v_w\}_{w \in \mathcal{D}}$, the trained vectors for words in $w \in \mathcal{D}$}
\end{itemize}

\begin{enumerate}[Step 1.]
\item Compute a rank-$k$ basis $U_k \in \mathbb{R}^{d \times k}$ for the subspace of category $c$ using Algorithm \ref{section:svd_basis}:
\[ U_k \leftarrow  \textsf{GET\_BASIS}( \{ v_w : w \in S_c \},\, k ).\]
Let $u_1$ be the first (column) basis vector of $U_k$.

\item Return the set of words $w \in \mathcal{D}\backslash S_{c}$ whose vectors have (i) a positive first coordinate $v_w \cdot u_1$ in the basis $U_k$, and (ii) a large enough projection $\len{ v_w U_{k} } > \delta$ in the subspace of category $c$:
\[
\{ w \in \mathcal{D}\backslash S_{c} : v_w \cdot u_1 > 0, \; \len{ v_w U_{k} } > \delta \}\,.
\]

\end{enumerate}

\noindent {\bf end function}
\end{framed}
\caption{$\textsf{EXTEND\_CATEGORY}( S_c, k, \delta)$ returns a set of words in $\mathcal{D} \backslash S_c$ which are likely to belong to category $c$.}
\label{section:discover_new_entities-algo}
\end{figure}
\renewcommand{\figurename}{Figure}
%%%%%%%%%%%%%%%%%%%%%%%%%%%%%%%%%%%%%%%%%%%%%%%%%%%%%%%%%%%%%%%%%%%%%%%%%%%%%%%%

\subsection{Performance}

We tested $\textsf{EXTEND\_CATEGORY}(S_c, k, \delta)$ on various categories $c$ from Figure \ref{fig:freebase_category}, using rank $k=10$ and threshold $\delta=0.6$. Table \ref{table:learn_new_entities} lists the top 5 words returned by the algorithm, where the words $w \in \mathcal{D} \backslash S_c$ are ordered in descending magnitude of the projection $\len{v_w U_k}$ onto the subspace $U_k$ of category $c$. The algorithm makes a few mistakes, e.g., it returns \texttt{london} as a \texttt{tourist\_attraction}, and \texttt{aren} as a \texttt{basketball\_player}. But overall, the algorithm seems to work very well and returns correct words that belong to the category.

\begin{table}\centering
\input{table-learn_new_facts-entities.tex}

\caption{Given a set $S_c \subset \mathcal{D}_c$ of words belonging to a category $c$, \textsf{EXTEND\_CATEGORY} (Algorithm \ref{section:discover_new_entities-algo}) returns new words in $\mathcal{D} \backslash S_c$ which are also likely to belong to $c$. Here, we list the top 5 words returned by $\textsf{EXTEND\_CATEGORY}(S_c, k, \delta)$ for various categories $c$ in Figure \ref{fig:freebase_category}, using rank $k=10$ and threshold $\delta=0.6$. The words $w \in \mathcal{D} \backslash S_c$ are ordered in descending magnitude of the projection $\len{v_w U_k}$ onto the category subspace. The algorithm makes a few mistakes, e.g., it returns \texttt{london} as a \texttt{tourist\_attraction}, and \texttt{aren} as a \texttt{basketball\_player}. But overall, the algorithm seems to work very well, and returns correct words that belong to the category.
\label{table:learn_new_entities}}
\end{table}

\section{Learning new word pairs in a relation} \label{section:learning_new_pairs_in_relation}

Let $r$ be a well-defined relation such that $\KB$ only knows a subset $S_r \subset \mathcal{D}_r$ of word pairs that satisfy $r$. We now present a method called \textsf{EXTEND\_RELATION} (Algorithm \ref{section:learn_new_relations-algo}) for discovering new word pairs in $(\mathcal{D} \times \mathcal{D})\backslash S_r$ that also satisfy $r$.

%%%%%%%%%%%%%%%%%%%%%%%%%%%%%%%%%%%%%%%%%%%%%%%%%%%%%%%%%%%%%%%%%%%%%%%%%%%%%%%%
\renewcommand{\figurename}{Algorithm}
\begin{figure}
\begin{framed}
\noindent {\bf function \textsf{EXTEND\_RELATION}($S_r$, $\{ k_A, k_B, k_r \}$, $\{ \delta_A, \delta_B, \delta_r \}$)}: Returns a set of word pairs in $(\mathcal{D} \times \mathcal{D}) \backslash S_r$.\\\\
\noindent Inputs:
\begin{itemize}
\item A subset $S_r \subset \mathcal{D}_r$ of word pairs that satisfy some well-defined relation $r$. Let $A := \{ a : (a,b) \in S_r \}$ and $B := \{ b : (a, b) \in S_r\}$. Let $c_A$ be the category such that, for all $a \in A$, $a$ belongs to category $c_A$. Similarly, let $c_B$ be the category such that, for all $b \in B$, $b$ belongs to category $c_B$.
\item $k_A, k_B, k_r \in \mathbb{N}$, the rank of the basis for the subspaces of $c_A$, $c_B$, and $r$ respectively
\item $\delta_A, \delta_B, \delta_r \in [0, 1]$, threshold values
\eat{\item $\{v_w\}_{w \in \mathcal{D}}$, the trained vectors for words in $w \in \mathcal{D}$}
\end{itemize}

\begin{enumerate}[Step 1.]
\item Use Algorithm \ref{section:discover_new_entities-algo} to get a set of words $S_A \subseteq \mathcal{D} \backslash A$ whose vectors have a large enough projection ($\geq \delta_A$) in the rank-$k_A$ subspace of category $c_A$, and a set of words $S_B \subseteq \mathcal{D} \backslash B$ whose vectors have a large enough projection ($\geq \delta_B$) in the rank-$k_B$ subspace of category $c_B$:
\begin{align*}
S_A &\leftarrow\textsf{EXTEND\_CATEGORY}( A, k_A, \delta_A)\\
S_B &\leftarrow\textsf{EXTEND\_CATEGORY}( B, k_B, \delta_B).
\end{align*}

\item Compute a rank-$k_r$ basis for the subspace of relation $r$ using Algorithm \ref{section:svd_basis}:
\[
U_{k_r} \leftarrow \textsf{GET\_BASIS}(\{ v_a - v_b : (a,b) \in S_r \},\, k_r).
\]
Let $u_1$ be the first (column) basis vector of $U_{k_r}$.

\item Return the set of word pairs $(a,b) \in S_A \times S_B$ whose difference vectors have (i) a positive first coordinate $(v_a-v_b) \cdot u_1$ in the basis $U_{k_r}$, and (ii) a large enough projection $\len{ (v_a - v_b) U_{k_r} } > \delta_r$ in the subspace of relation $r$:
\[
\{ (a,b) \in S_A \times S_B : (v_a-v_b) \cdot u_1 > 0, \; \len{ (v_a-v_b) U_k } > \delta_r \} \,.
\]

\end{enumerate}

\noindent {\bf end function}
\end{framed}
\caption{\textsf{EXTEND\_RELATION}($S_r$, $\{ k_A, k_B, k_r \}$, $\{ \delta_A, \delta_B, \delta_r \}$) returns a set of word pairs $(a,b) \in (\mathcal{D} \times \mathcal{D}) \backslash S_r$ which are likely to satisfy relation $r$.}
\label{section:learn_new_relations-algo}
\end{figure}
\renewcommand{\figurename}{Figure}
%%%%%%%%%%%%%%%%%%%%%%%%%%%%%%%%%%%%%%%%%%%%%%%%%%%%%%%%%%%%%%%%%%%%%%%%%%%%%%%%

%%%%%%%%%%%%%%%%%%%%%%%%%%%%%%%%%%%%%%%%%%%%%%%%%%%%%%%%%%%%%%%%%%%%%%%%%%%%%
\subsection{Experiment}\label{section:experiment-learn_new_facts}

We tested \textsf{EXTEND\_RELATION} on three well-defined relations from Figure \ref{fig:freebase_relation}: \texttt{capital\_city}, \texttt{city\_in\_state}, and \texttt{currency\_used}. For each of the relations $r$, let $S_r \subseteq \mathcal{D}_r$ be the set of word pairs contained in the corresponding relation file (see Figure \ref{fig:freebase_relation}(f)-(h)). We used cross-validation to assess the accuracy rate of Algorithm \ref{section:learn_new_relations-algo}, as follows. For each rank $k \in \{1, 2, \ldots, 9\}$ and threshold $\delta \in \{ 0.4, 0.45, \ldots,  0.75 \}$, we repeated the following experiment over $T=50$ trials:

\begin{enumerate}[Step 1.]
\item Randomly partition $S_r$ into a training set $S_1$ and a testing set $S_2$, where the training set size is $|S_1| = 0.3|S_r|$. Let $A := \{ a : (a,b) \in S_2\}$ and $B := \{ b : (a,b) \in S_2\}$.

\item Use Algorithm \ref{section:learn_new_relations-algo} to find a set $\mathcal{S}$ of word pairs in $(\mathcal{D}\times \mathcal{D}) \backslash S_1$ which are likely to satisfy relation $r$:
\[
\mathcal{S} \leftarrow \textsf{EXTEND\_RELATION}( S_1, \{ k,k,k\}, \{ \delta, \delta, \delta \}).
\]

\item Let $\mathcal{S}' := \{ (a,b) \in \mathcal{S} : \textrm{$a \in A$ or $b \in B$} \}$.\footnote{We ignore the answers $(a,b) \in \mathcal{S}$ such that $a \notin A$ and $b \notin B$, because we do not have an automated way of determining whether it is correct or incorrect. One could check each of these answers manually using an external knowledge source (e.g., Google search), but doing so would be very time-consuming.} For each answer $(a,b) \in \mathcal{S}'$, we count it as correct if $(a,b) \in S_2$, and incorrect otherwise. So the resulting accuracy of \textsf{EXTEND\_RELATION} using parameters $k$ and $\delta$ is
\[
\acc(k,\delta) := \frac{\textrm{\# correct answers}}{\textrm{\# total answers}}= \frac{\left|\mathcal{S}' \cap S_2\right|}{|\mathcal{S}'|}\,.
\]
\end{enumerate}

%%%%%%%%%%%%%%%%%%%%%%%%%%%%%%%%%%%%%%%%%%%%%%%%%%%%%%%%%%%%%%%%%%%%%%%%%%%%%
\subsection{Performance}

\begin{figure}\centering
\begin{subfigure}{\textwidth}
\includegraphics[width=\textwidth]{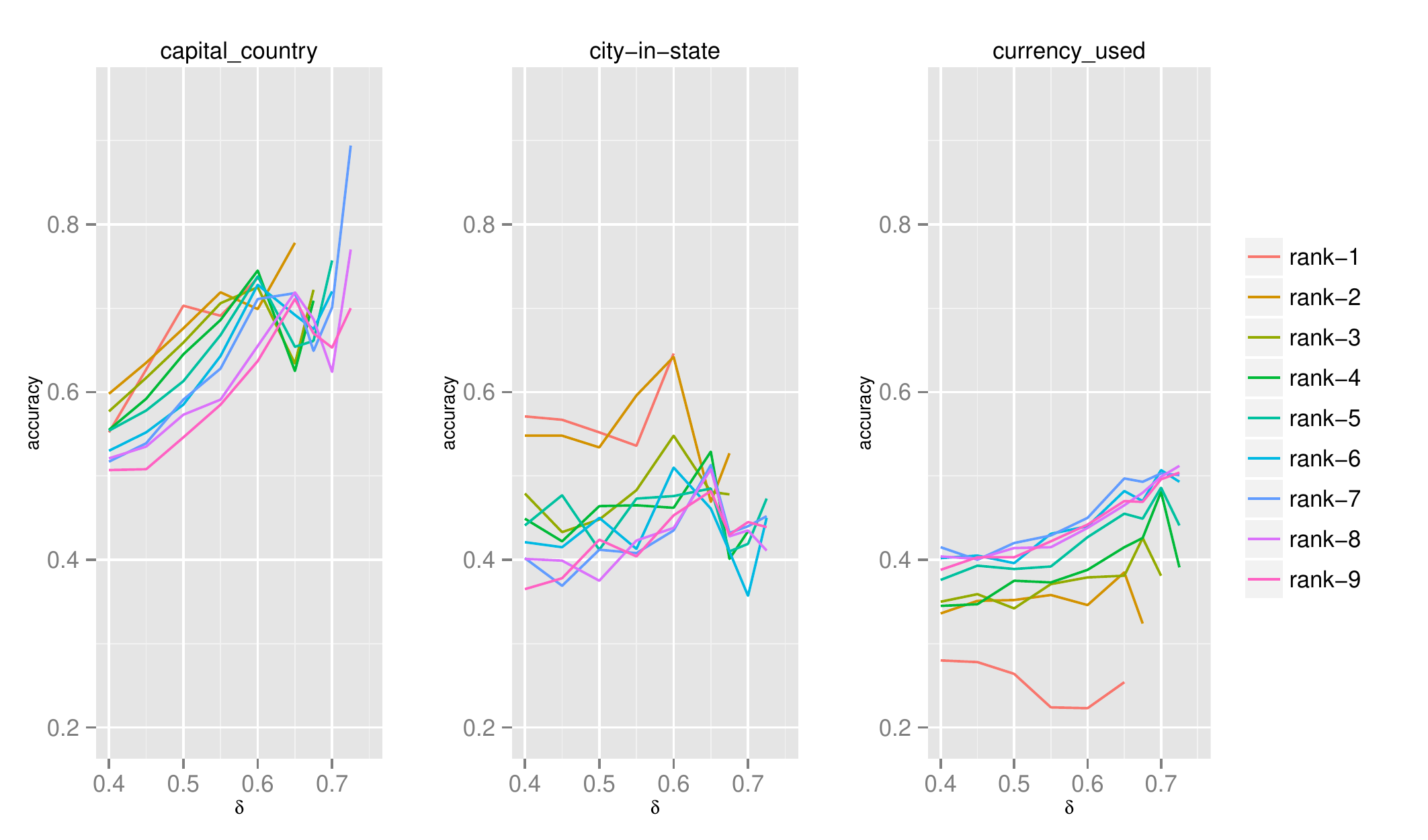}
\caption{For each rank $k \in \{1, 2, \ldots, 9\}$ and threshold $\delta \in \{ 0.4, 0.45, \ldots,  0.75 \}$, we plot the average accuracy $\frac{1}{T} \sum_{t=1}^T \acc^{(t)}(k, \delta)$ over $T=50$ trials, where for each trial $t$, $\acc^{(t)}(k,\delta) = \frac{\textrm{\# correct answers}}{\textrm{\# total answers}}$ is the accuracy of the answers returned by $\textsf{EXTEND\_RELATION}(S_1^{(t)}, \{k,k,k \}, \{\delta,\delta,\delta \})$. (Here, each $S_1^{(t)} \subset \mathcal{D}_r$ is a random subset of the word pairs contained in the relation file of $r$ (see Figure \ref{fig:freebase_relation}(f)-(h)). $S_1^{(t)}$ is generated randomly, at each trial $t$, in Step 1 of the experiment in Section \ref{section:experiment-learn_new_facts}. ). \label{fig:discover_new_facts-acc_all}}
\end{subfigure}

\vspace{5ex}
\begin{subfigure}{\textwidth}\centering
\begin{tabular}{| t || c | c | c |}
\hline
$r$ & $\max_{k, \delta} \acc(k, \delta) $ & rank $k$ & threshold $\delta$ \\ \hline
\texttt{capital\_city} & 0.909 & 7 & 0.75 \\
\texttt{city\_in\_state} & 0.646 & 1 & 0.6\\
\texttt{currency\_used} & 0.507 & 6 & 0.7\\
\hline
\end{tabular}
\caption{For each relation $r$, we list the parameters $k$ and $\delta$ that achieved the highest average accuracy in (a).\label{table:discover_new_facts-best_acc}}
\end{subfigure}
\caption{Given a set $S_r \subset \mathcal{D}_r$ of word pairs satisfying a relation $r$, \textsf{EXTEND\_RELATION} (Algorithm \ref{section:learn_new_relations-algo}) returns new word pairs in $(\mathcal{D} \times \mathcal{D})\backslash S_r$ which are also likely to satisfy relation $r$. We use cross-validation to assess the accuracy rate of \textsf{EXTEND\_RELATION}  on three well-defined relations from Figure \ref{fig:freebase_relation}(f)-(h).
Note that \textsf{EXTEND\_RELATION} performs very well on $r=\texttt{capital\_city}$, achieving accuracy as high as $0.909$. We discuss why the accuracy is lower for the other two relations in Section \ref{section:extend_relations-explanation}. For details about the experiment, see Section \ref{section:experiment-learn_new_facts}.}
\label{fig:discover_new_facts-performance}
\end{figure}

For each rank $k \in \{1, 2, \ldots, 9\}$ and threshold $\delta \in \{ 0.4, 0.45, 0.5, \ldots,  0.75 \}$, we plot the average accuracy $\frac{1}{T} \sum_{t=1}^T \acc^{(t)}(k,\delta)$ over $T=50$ trials\footnote{where $\acc^{(t)}(k,\delta)$ is the accuracy obtained in trial $t$.} in Figure \ref{fig:discover_new_facts-acc_all}. The table in Figure \ref{table:discover_new_facts-best_acc} lists the parameters $k$ and $\delta$ that attained the highest average accuracy. The algorithm achieved significantly higher accuracy for $r=\texttt{capital\_city}$ than for the other two relations; we discuss why in Section \ref{section:extend_relations-explanation}. However, the performance of \textsf{EXTEND\_RELATION} is still quite remarkable, considering the fact that it returns reasonable word pairs out of the $\binom{|\mathcal{D}|}{2} \approx 1.8\textnormal{e}\!+\!09$ possible word pairs in $\mathcal{D} \times \mathcal{D}$.

\begin{figure}\centering
\includegraphics[width=\textwidth]{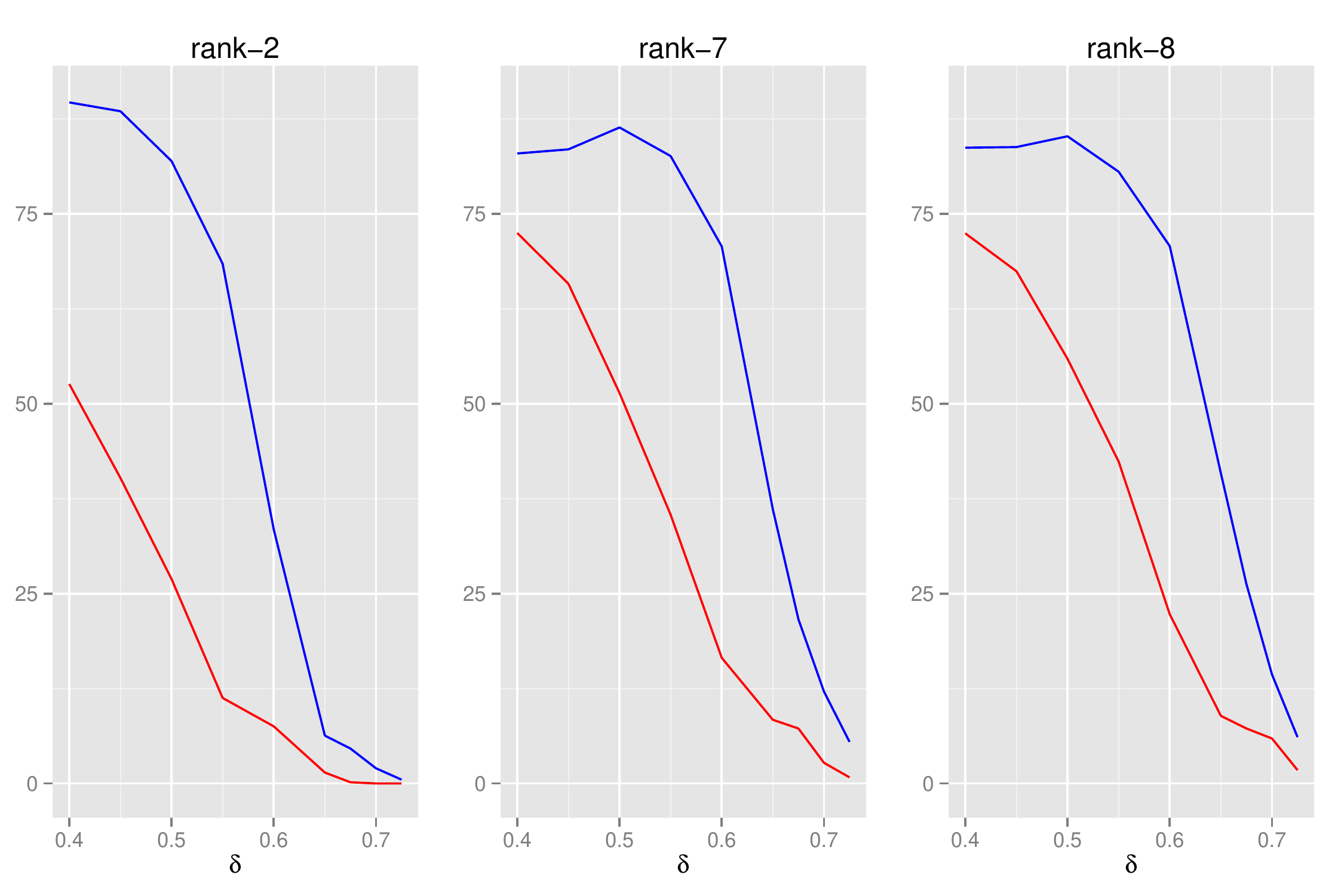}
\caption{Let $r=\texttt{capital\_city}$, and let $S_1 \subset \mathcal{D}_r$ be a random subset of the word pairs contained in the relation file of $r$ (see Figure \ref{fig:freebase_relation}(g)). We plot the number of correct ({\color{blue}blue}) and incorrect ({\color{red}red}) word pairs returned, respectively, by calling $\textsf{EXTEND\_RELATION}(S_1,\, \{ k,k,k \}, \{ \delta,\, \delta,\, \delta \})$ for varying threshold values $\delta$ (see Algorithm \ref{section:learn_new_relations-algo}). We only provide plots for the ranks $k \in \{2, 7, 8\}$; the plots for other ranks are similar. As the threshold $\delta$ is increased, the algorithm filters out more word pairs in Steps 2 and 3 of the algorithm, resulting in a smaller number of word pairs returned by the algorithm. \label{fig:discover_new_facts-capital_country-numcorrect}}
\end{figure}

As the threshold $\delta$ is increased, the algorithm filters out more words in Step 2 and word pairs in Step 3 of the algorithm, resulting in a smaller number of word pairs returned by the algorithm. Figure \ref{fig:discover_new_facts-capital_country-numcorrect} illustrates this effect for $r=\texttt{capital\_city}$.

Note that the algorithm's accuracy can be further improved by fine-tuning the parameters. In our experiment (Section \ref{section:experiment-learn_new_facts}), we used the same rank $k$ for $k_A$, $k_B$, and $k_r$, and also used the same threshold $\delta$ for $\delta_A$, $\delta_B$, and $\delta_r$; but one can vary each of these parameters separately to achieve better performance.

\subsection{Varying levels of difficulty for different relations}\label{section:extend_relations-explanation}

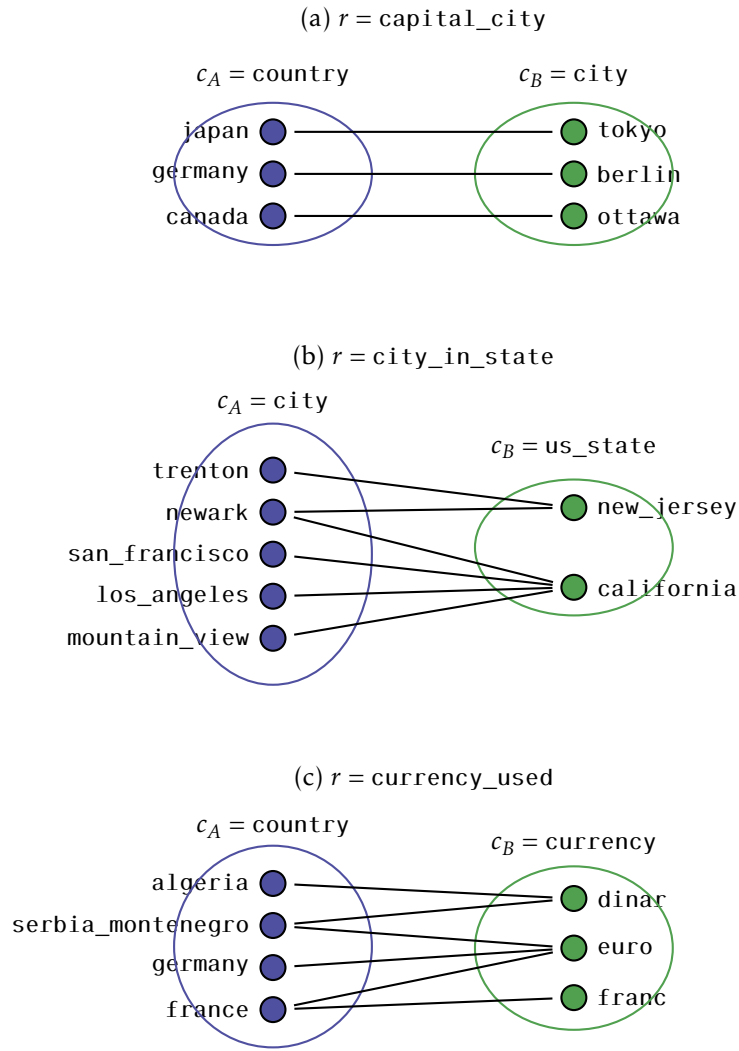
\begin{figure}\centering
\input{tikz2.tex}
\caption{For $r=\texttt{capital\_city}$, there is a one-to-one mapping between $A_r := \{ a : (a,b) \in S_r \}$ and $B_r := \{ b : (a,b) \in S_r \}$, since a country has exactly one capital city. The same does not hold for (b) or (c), however: (b) Each U.S. state contains multiple cities, and some cities in different U.S. states have identical names (e.g., both California and New Jersey have a city named Newark). (c) A country can have multiple currencies (either concurrently or over history), and the same currency can be used in multiple countries. This causes \textsf{EXTEND\_RELATION} to return a higher number of false-positive (incorrect) answers for (b) and (c); see Section \ref{section:extend_relations-explanation} for a detailed explanation. \label{fig:relation_types}}
\end{figure}

We provide two explanations as to why \textsf{EXTEND\_RELATION} underperforms on relations such as $\texttt{city\_in\_state}$ and  $\texttt{currency\_used}$:
\begin{enumerate}

\item For $r=\texttt{capital\_city}$, there is a one-to-one mapping\footnote{In other words, given a word $a \in A_r$, there exists a unique word $b \in B_r$ such that $(a,b)$ satisfies the relation $r$; and conversely, given a word $b \in B_r$, there exists a unique word $a \in A_r$ such that $(a,b)$ satisfies $r$.} between the sets $A_r := \{ a : (a,b) \in S_r \}$ and $B_r := \{ b : (a,b) \in S_r \}$, whereas the same does not hold for $r=\texttt{city\_in\_state}$ or $r=\texttt{currency\_used}$ (see Figure \ref{fig:relation_types}). This causes the algorithm to return many false-positive answers for the relations $\texttt{city\_in\_state}$ and $\texttt{currency\_used}$, as we illustrate with an example below.

Consider the relation $r=\texttt{currency\_used}$. In the set $S_r$ of word pairs contained in the relation file for $r$ (see Figure \ref{fig:freebase_relation}(h)), there are 30 country-currency pairs $(a, b) \in S_r$ where $b=\texttt{franc}$. For 24 of these pairs $\{ (a, \texttt{franc}) \in S_r \}$, the country word $a$ belongs to the category $c=\texttt{african\_country}$. Because the low-rank basis $U_{k_r}$ for relation $r$ (computed in Step 2 of Algorithm \ref{section:learn_new_relations-algo}) tries to capture the vectors in the set $\{ v_a-v_{\texttt{franc}} : (a,\,\texttt{franc}) \in S_r \}$, and the vectors of words belonging to the category $c=\texttt{african\_country}$ are clustered together, the algorithm returns many false-positive pairs consisting of an African country and the currency \texttt{franc}. For example, in many trial runs, the algorithm returns incorrect pairs such as (\texttt{kenya}, \texttt{franc}), (\texttt{uganda}, \texttt{franc}), and (\texttt{sudan}, \texttt{franc}). False-positive answers such as these cause the algorithm's accuracy to drop.

\item Some relations are just inherently more difficult than others to represent using word vectors. For example, Figure \ref{fig:wordnet_filter-facts} shows that solving analogy queries of the form ``a:b::c:??'' for pairs $(a,b), (c,d)$ in the relation file for \texttt{country-currency} is more difficult than for pairs $(a,b), (c,d)$ satisfying the relation \texttt{country-capital2}. This may explain why \textsf{EXTEND\_RELATION} performs worse on $\texttt{currency\_used}$ than on \texttt{capital\_city}.\footnote{\texttt{country-currency} is a smaller subset of the relation file for $\texttt{currency\_used}$, and \texttt{country-capital2} is a smaller subset of \texttt{capital\_city}; see Figure \ref{fig:freebase_relation}.}
\end{enumerate}

\section{Conclusion}

We have demonstrated that the low-dimensional subspace of categories and relations can be used to discover new facts with fairly low false-positive rates. The performance of \textsf{EXTEND\_RELATION} (Algorithm \ref{section:learn_new_relations-algo}) is especially surprising, given the simplicity of the algorithm and the fact that it returns plausible word pairs out of all possible word pairs in $\mathcal{D} \times \mathcal{D}$. The algorithms \textsf{EXTEND\_CATEGORY} (Algorithm \ref{section:discover_new_entities-algo}) and \textsf{EXTEND\_RELATION} (Algorithm \ref{section:learn_new_relations-algo}) are computationally efficient, are shown to drastically narrow down the search space for discovering new facts, and can be used to supplement other methods for extending knowledge bases.

%%%%%%%%%%%%%%%%%%%%%%%%%%%%%%%%%%%%%%%%%%%%%%%%%%%%%%%%%%%%%%%%%%%%%%%%%%%%%
%%%%%%%%%%%%%%%%%%%%%%%%%%%%%%%%%%%%%%%%%%%%%%%%%%%%%%%%%%%%%%%%%%%%%%%%%%%%%

%%%%%%%%%%%%%%%%%%%%%%%%%%%%%%%%%%%%%%%%%%%%%%%%%%%%%%%%%%%%%%%%%%%%%%%%%%%%%
%%%%%%%%%%%%%%%%%%%%%%%%%%%%%%%%%%%%%%%%%%%%%%%%%%%%%%%%%%%%%%%%%%%%%%%%%%%%%

%%%%%%%%%%%%%%%%%%%%%%%%%%%%%%%%%%%%%%%%%%%%%%%%%%%%%%%%%%%%%%%%%%%%%%%%%%%%%
%%%%%%%%%%%%%%%%%%%%%%%%%%%%%%%%%%%%%%%%%%%%%%%%%%%%%%%%%%%%%%%%%%%%%%%%%%%%%

%% file: table-learn_new_facts-entities.tex
\begin{subtable}{.5\textwidth}\centering
\subcaption{$c=\texttt{classical\_composer}$}
\begin{tabular}{ | t | c |}\hline
word & projection\\ \hline
schumann & 0.841 \\
beethoven & 0.840 \\
stravinsky & 0.796 \\
liszt & 0.789 \\
schubert & 0.788 \\
\hline\end{tabular}
\end{subtable}\begin{subtable}{.5\textwidth}\centering\subcaption{$c=\texttt{sport}$}\begin{tabular}{ | t | c |}\hline
word & projection\\ \hline
biking & 0.881 \\
volleyball & 0.870 \\
skiing & 0.810 \\
softball & 0.809 \\
soccer & 0.801 \\
\hline\end{tabular}\end{subtable}\vspace{5ex}

\begin{subtable}{.5\textwidth}\centering\subcaption{$c=\texttt{university}$}\begin{tabular}{ | t | c |}\hline
word & projection\\ \hline
cambridge\_university & 0.897 \\
university\_of\_california & 0.889 \\
new\_york\_university & 0.868 \\
stanford\_university & 0.824 \\
yale\_university & 0.822 \\
\hline\end{tabular}\end{subtable}\vspace{5ex}\begin{subtable}{.5\textwidth}\centering\subcaption{$c=\texttt{basketball\_player}$}\begin{tabular}{ | t | c |}\hline
word & projection\\ \hline
dwyane\_wade & 0.788 \\
aren & 0.721 \\
kobe\_bryant & 0.715 \\
chris\_bosh & 0.712 \\
tim\_duncan & 0.708 \\
\hline\end{tabular}\end{subtable}

\begin{subtable}{.5\textwidth}\centering\subcaption{$c=\texttt{religion}$}\begin{tabular}{ | t | c |}\hline
word & projection\\ \hline
christianity & 0.899 \\
hinduism & 0.880 \\
taoism & 0.863 \\
buddhist & 0.846 \\
judaism & 0.830 \\
\hline\end{tabular}\end{subtable}\begin{subtable}{.5\textwidth}\centering\subcaption{$c=\texttt{tourist\_attraction}$}\begin{tabular}{ | t | c |}\hline
word & projection\\ \hline
metropolitan\_museum\_of\_art & 0.822 \\
museum\_of\_modern\_art & 0.813 \\
london & 0.764 \\
national\_gallery & 0.764 \\
tate\_gallery & 0.756 \\
\hline\end{tabular}\end{subtable}\vspace{5ex}

\begin{subtable}{.5\textwidth}\centering\subcaption{$c=\texttt{holiday}$}\begin{tabular}{ | t | c |}\hline
word & projection\\ \hline
diwali & 0.821 \\
christmas & 0.806 \\
passover & 0.784 \\
new\_year & 0.783 \\
rosh\_hashanah & 0.749 \\
\hline\end{tabular}\end{subtable}\begin{subtable}{.5\textwidth}\centering\subcaption{$c=\texttt{month}$}\begin{tabular}{ | t | c |}\hline
word & projection\\ \hline
august & 0.988 \\
april & 0.987 \\
october & 0.985 \\
february & 0.983 \\
november & 0.980 \\
\hline\end{tabular}\end{subtable}\vspace{5ex}

\begin{subtable}{.5\textwidth}\centering\subcaption{$c=\texttt{animal}$}\begin{tabular}{ | t | c |}\hline
word & projection\\ \hline
horses & 0.806 \\
moose & 0.784 \\
elk & 0.783 \\
raccoon & 0.763 \\
goats & 0.762 \\
\hline\end{tabular}\end{subtable}\begin{subtable}{.5\textwidth}\centering\subcaption{$c=\texttt{asian\_city}$}\begin{tabular}{ | t | c |}\hline
word & projection\\ \hline
taipei & 0.837 \\
taichung & 0.819 \\
kaohsiung & 0.818 \\
osaka & 0.806 \\
tianjin & 0.765 \\
\hline\end{tabular}\end{subtable}

%% file: tikz2.tex
\definecolor{myblue}{RGB}{80,80,160}
\definecolor{mygreen}{RGB}{80,160,80}

\begin{tikzpicture}[thick,
  %every node/.style={draw,circle},
  fsnode/.style={draw, circle,fill=myblue},
  ssnode/.style={draw,circle,fill=mygreen},
  %emptynode/.style={empty},
  every fit/.style={ellipse,draw,inner sep=-2pt,text width=2cm},
  -,shorten >= 3pt,shorten <= 3pt
]

\node[yshift=1.5cm ] {(a) $r=\texttt{capital\_city}$};
\node[yshift=0.75cm, xshift=-2cm ] {$c_A=\texttt{country}$};
\node[yshift=0.75cm, xshift=2cm ] {$c_B=\texttt{city}$};

% the vertices of U
\begin{scope}[xshift=-2cm,start chain=going below,node distance=2mm, align=center]
\node[fsnode, on chain] (f1) [label=left: \texttt{japan}] {};
\node[fsnode, on chain] (f2) [label=left: \texttt{germany}] {};
\node[fsnode, on chain] (f3) [label=left: \texttt{canada}] {};
\end{scope}

% the vertices of V
\begin{scope}[xshift=2cm,start chain=going below,node distance=2mm]
\node[ssnode,on chain] (s1) [label=right: \texttt{tokyo}] {};
\node[ssnode,on chain] (s2) [label=right: \texttt{berlin}] {};
\node[ssnode,on chain] (s3) [label=right: \texttt{ottawa}] {};
\end{scope}

\node [myblue,fit=(f1) (f3)]{};%, label=left:$\textrm{(I)}\qquad\qquad\qquad\qquad$]{};%, label=above:\texttt{country}] {};
\node [mygreen,fit=(s1) (s3)]{};%,label=above:\texttt{city}] {};

% the edges
\draw (f1) -- (s1);
\draw (f2) -- (s2);
\draw (f3) -- (s3);

%%%%%%%%%%%%%%%%%%%%%%%%%%%%%%%%%%%%%%%%%%%%%%%%%%%%%%%%%%%%%%%%%%%%%
% the vertices of U

\node[yshift=-3cm ]{(b) $r=\texttt{city\_in\_state}$};
\node[yshift=-3.6cm, xshift=-2cm ] {$c_A=\texttt{city}$};
\node[yshift=-4.2cm, xshift=2cm ] {$c_B=\texttt{us\_state}$};

\begin{scope}[yshift=-4.5cm,xshift=-2cm,start chain=going below,node distance=2mm]
\node[fsnode, on chain] (a1) [label=left: \texttt{trenton}] {};
\node[fsnode, on chain] (a2) [label=left: \texttt{newark}] {};
\node[fsnode, on chain] (a3) [label=left: \texttt{san\_francisco}] {};
\node[fsnode, on chain] (a4) [label=left: \texttt{los\_angeles}] {};
\node[fsnode, on chain] (a5) [label=left: \texttt{mountain\_view}] {};
\end{scope}

% the vertices of V
\begin{scope}[xshift=2cm, yshift=-5cm, start chain=going below,node distance=7mm]
\node[ssnode,on chain] (b1) [label=right: \texttt{new\_jersey}] {};
\node[ssnode,on chain] (b2) [label=right: \texttt{california}] {};
\end{scope}

\node [myblue,fit=(a1) (a5)]{};%, label=left:$\textrm{(II)}\qquad\qquad\qquad\qquad$]{};
\node [mygreen,fit=(b1) (b2)]{};%,label=above:\texttt{us\_state}] {};

% the edges
\draw (a1) -- (b1);
\draw (a2) -- (b1);
\draw (a2) -- (b2);
\draw (a3) -- (b2);
\draw (a4) -- (b2);
\draw (a5) -- (b2);
%%%%%%%%%%%%%%%%%%%%%%%%%%%%%%%%%%%%%%%%%%%%%%%%%%%%%%%%%%%%%%%%%%%%%

\node[yshift=-8.6cm ] {(c) $r=\texttt{currency\_used}$};
\node[yshift=-9.25cm , xshift=-2cm] {$c_A=\texttt{country}$};
\node[yshift=-9.5cm , xshift=2cm] {$c_B=\texttt{currency}$};
% the vertices of U
\begin{scope}[yshift=-10cm,xshift=-2cm,start chain=going below,node distance=2mm, align=center]
\node[fsnode, on chain] (c1) [label=left: \texttt{algeria}] {};
\node[fsnode, on chain] (c2) [label=left: \texttt{serbia\_montenegro}] {};
\node[fsnode, on chain] (c3) [label=left: \texttt{germany}] {};
\node[fsnode, on chain] (c4) [label=left: \texttt{france}] {};
\end{scope}

% the vertices of V
\begin{scope}[xshift=2cm, yshift=-10.2cm, start chain=going below,node distance=3mm]
\node[ssnode,on chain] (d1) [label=right: \texttt{dinar}] {};
\node[ssnode,on chain] (d2) [label=right: \texttt{euro}] {};
\node[ssnode,on chain] (d3) [label=right: \texttt{franc}]{};
\end{scope}

\node [myblue,fit=(c1) (c4)]{};%, label=left:$\textrm{(III)}\qquad\qquad\qquad\qquad$]{};
\node [mygreen,fit=(d1) (d3)]{};%, label=right:$\qquad\qquad\qquad\qquad$]{};%]{};%,label=above:\texttt{currency}] {};

% the edges
\draw [label=right:\texttt{currency\_used}]  (c1) -- (d1);
\draw (c2) -- (d1);
\draw (c2) -- (d2);
\draw (c3) -- (d2);
\draw (c4) -- (d2);
\draw (c4) -- (d3);
\end{tikzpicture}

%% file: fit_new_vec.tex
\chapter{Learning vectors for less frequent words}\label{chapter:fit_new_vec}
\chaptermark{Learning vectors}
Since the size of the co-occurrence data is quadratic in the size of the vocabulary $\mathcal{D}$, and since the co-occurrence data for infrequent words are too noisy to generate good word vectors with, we restrict the vocabulary $\mathcal{D}$ to only the words that appear at least $m_0 = 1000$ times in the corpus. Hence, we do not have vectors for words that appear fewer than $1000$ times in the corpus.\footnote{There are a total of $283,847$ words that occur between 100 and 999 times in the corpus, which are not included in $\mathcal{D}$.} These words include the famous composer \texttt{claude\_debussy} (994 times), Malaysian currency \texttt{ringgit} (369 times), the famous actor \texttt{adam\_sandler} (982 times), and the historical event \texttt{boston\_massacre} (342 times). In order to continually extend the knowledge base $\KB$, it becomes necessary to learn vectors for these less frequent words.

We demonstrate that, using the low-dimensional subspace of categories, one can substantially reduce the amount of co-occurrence data needed to learn vectors for words. In particular, we present an algorithm called \textsf{LEARN\_VECTOR} (Algorithm \ref{section:fit_new_vec-algo}) for learning vectors of words with only a small amount of co-occurrence data. We test the algorithm on seven words $\hat{w}$ (listed in Table \ref{table:fit_new_vec-words}) which we already have vectors for, and compare the ``true'' vector $v_{\hat{w}} \in V_{\mathcal{D}}$ to the vector $\hat{v}$ returned by \textsf{LEARN\_VECTOR}. The algorithm's performance is given in Figure \ref{fig:fit_new_vec}. In general, the algorithm achieves very good performance while using only a small fraction of the total amount of co-occurrence data in the Wikipedia corpus $\mathcal{C}$.

One can extend this method to learn vectors for \emph{any} words -- even words that do not appear at all in the Wikipedia corpus -- using web-scraping tools, such as Google search, to obtain additional co-occurrence data.

\section{Learning a new vector}

Let $\hat{w}$ be a word such that (i) we know the category $c$ that $\hat{w}$ belongs in, and (ii) we have a small corpus $\Gamma$ (where $|\Gamma| \ll |\mathcal{C}|$) containing co-occurrence data for $\hat{w}$. Then we provide a method \textsf{LEARN\_VECTOR} for learning its word vector $\hat{v} \in \mathbb{R}^d$ (see Algorithm \ref{section:fit_new_vec-algo}).

%%%%%%%%%%%%%%%%%%%%%%%%%%%%%%%%%%%%%%%%%%%%%%%%%%%%%%%%%%%%%%%%%%%%%%%%%%%%%%%%
\renewcommand{\figurename}{Algorithm}
\begin{figure}
\begin{framed}
\noindent {\bf function $\textsf{LEARN\_VECTOR}( \hat{w}, c, \Gamma, k, \eta, \lambda)$}: Returns a learned vector $\hat{v} \in \mathbb{R}^d$ for word $\hat{w}$.\\\\
\noindent Inputs:
\begin{itemize}
\item $\hat{w}$, a word
\item $c$, the category that $\hat{w}$ belongs in. Let $V_c := \{ v_w : w \in \mathcal{D}_c \}$ be the vectors of words belonging to $c$.
\item $\Gamma$, a small corpus containing co-occurrence data for $\hat{w}$
\item $k \in \mathbb{N}$, the rank of the basis for the category subspace
\item $\eta \in (0, 1]$, the learning rate for Adagrad
\item $\lambda > 0$, the weight of the regularization term in the objective (\ref{eq:objective-fit_new_vec})
\eat{\item $\{ v_w \}_{w \in \mathcal{D}}$, the trained vectors for words in $\mathcal{D}$}
\end{itemize}

\begin{enumerate}[Step 1.]
\item Compute a rank-$k$ basis $U_k \in \mathbb{R}^{d \times k}$ for the subspace of category $c$ using Algorithm \ref{section:svd_basis}:
\[ U_k \leftarrow  \textsf{GET\_BASIS}( V_c,\, k ).\]

\item Consider only the set $D'$ of vocabulary words that appear at least $m_0'=10$ times in $\Gamma$. For each word $w \in D'$, compute $Y_{\hat{w}w}$, the number of times word $w$ appears in any context window around $\hat{w}$ in $\Gamma$.

\item Use Adagrad with learning rate $\eta$ to train parameters $\hat{v} \in \mathbb{R}^d$, $b \in \mathbb{R}^k$, and $Z \in \mathbb{R}$ so as to minimize the objective
\begin{equation}
\sum_{w \in D' \cap \mathcal{D}} g(Y_{\hat{w}w}) \paren*{||\hat{v} +v_{w}||^2 - \log(Y_{\hat{w}w}) +Z}^2 + \lambda \len{\hat{v} - U_k b}^2,
\label{eq:objective-fit_new_vec}\end{equation}
where $\{ v_{w} \}_{w \in \mathcal{D}}$ are the already-learned word vectors, and  $g(x) := \min\left\{ \paren*{\frac{x}{10}}^{0.75}, 1 \right\}$. Note that $\hat{v}$, $b$, and $Z$ are initialized randomly.

\item Normalize $\hat{v}$ so that $\len{\hat{v}} = 1$.

\item Return $\hat{v}$, the learned word vector for $\hat{w}$.
\end{enumerate}

\noindent {\bf end function}
\end{framed}
\caption{$\textsf{LEARN\_VECTOR}(\hat{w}, c, \Gamma, k, \eta, \lambda)$ returns a vector $\hat{v} \in \mathbb{R}^d$ for word $\hat{w}$ learned using the objective (\ref{eq:objective-fit_new_vec}).}
\label{section:fit_new_vec-algo}
\end{figure}
\renewcommand{\figurename}{Figure}
%%%%%%%%%%%%%%%%%%%%%%%%%%%%%%%%%%%%%%%%%%%%%%%%%%%%%%%%%%%%%%%%%%%%%%%%%%%%%%%%

\subsection{Motivation behind the optimization objective}

The objective (\ref{eq:objective-fit_new_vec}) is nearly identical to the Squared Norm (SN) objective (\ref{eq:objective-sq}), except for a few differences: (i) we have an additional regularization term $\lambda \len{\hat{v} - U_kb}^2$, (ii) the co-occurrence counts $Y_{\hat{w}w}$ are taken from the smaller corpus $\Gamma$, and (iii) the summation is taken over words $w \in D' \cap \mathcal{D}$. Note that $b$ has a closed-form solution, since to minimize $\len{\hat{v} - U_kb}^2$, one can just take $b$ to be the projection of $\hat{v}$ onto $U_k$. The regularization term $\lambda \len{\hat{v} - U_kb}^2$ serves as a prior knowledge, forcing the new vector $\hat{v}$ to be trained \emph{near} the subspace $U_k$ of category $c$, but also allowing $\hat{v}$ to lie \emph{outside} the subspace. The hope is that the regularization term reduces the amount of co-occurrence data needed to fit $\hat{v}$. Note that in general, the regularization weight $\lambda$ should be decreasing in the size of $\Gamma$, since less prior knowledge is needed with more data.

\section{Experiment}\label{section:experiment-fit_new_vec}

For our experiment, we chose seven words $\hat{w}$ (listed in Table \ref{table:fit_new_vec-words}) which we already have vectors for, fitted a vector $\hat{v}$ using \textsf{LEARN\_VECTOR} (Algorithm \ref{section:fit_new_vec-algo}), and compared $\hat{v}$ to the true vector $v_{\hat{w}} \in V_{\mathcal{D}}$ (see Figure \ref{fig:fit_new_vec}). We withheld the true vector $v_{\hat{w}} \in V_{\mathcal{D}}$ from training, by taking $\hat{w}$ out of the summation over $D' \cap \mathcal{D}$ in the objective (\ref{eq:objective-fit_new_vec}). For the words \texttt{california}, \texttt{germany}, and \texttt{japan}, we used $k=10$ as the rank of the basis of the category subspace; for the remaining four words, we used rank $k=3$.

For each word $\hat{w}$, we use $Y_{\hat{w}}(\Gamma) := \sum_{w \in D'} Y_{\hat{w}w}$ to quantify the amount of co-occurrence data for word $\hat{w}$ in a corpus $\Gamma$ with vocabulary set $D'$, and evaluate the algorithm in Section \ref{section:fit_new_vec-algo} for varying values of $Y_{\hat{w}}(\Gamma)$. More specifically, we extracted six subcorpora $\Gamma_1, \ldots, \Gamma_6$ from the original corpus $\mathcal{C}$, where $\Gamma_1 \subset \Gamma_2 \subset \cdots \subset \Gamma_6 \subset \mathcal{C}$ and $Y_{\hat{w}}(\Gamma_1) < Y_{\hat{w}}(\Gamma_2) < \cdots < Y_{\hat{w}}(\Gamma_6) \ll Y_{\hat{w}}(\mathcal{C}) = X_{\hat{w}}$. For each $i \in \{1, \ldots, 6\}$, we ran the algorithm with various learning rates $\eta_i$ and regularization weights $\lambda_i$; Table \ref{table:fit_new_vec-param} lists the parameter values that resulted in the best performance for each corpus size and each word.

%%%%%%%%%%%%%%%%%%%%%%%%%%%%%%%%%%%%%%%%%%%%%%%%%%%%%%%%%%%%%%%%%%%%%%%%%%%%%%%%%%%%%%%%%
\begin{table}\centering
  \begin{tabular}{ | t || t | c | c |}
    \hline
    $\hat{w}$ & $c$ & $k$ & $X_{\hat{w}}$   \\ \hline
    california  & us\_state & 10 & 4.094e+06 \\
    christianity  & religion  & 3 & 4.173e+05\\
    germany & country  & 10 & 3.082e+06\\
    hinduism & religion & 3  & 1.037e+05\\
    japan & country & 10 & 2.729e+06 \\
    massachusetts  & us\_state & 3 & 1.312e+06\\
    princeton  & university & 3 & 2.829e+05\\
    \hline
  \end{tabular}
\caption{In the experiment, we train vectors for the words $\hat{w}$ by minimizing the objective (\ref{eq:objective-fit_new_vec}). $c$ is the category that $\hat{w}$ belongs in. For the words \texttt{california}, \texttt{germany}, and \texttt{japan}, we used $k=10$ as the rank of the basis of the category subspace; for the remaining four words, we used rank $k=3$. $X_{\hat{w}} := \sum_{w \in \mathcal{C}}X_{\hat{w}w}$ measures the amount of co-occurrence data for $\hat{w}$ in the original corpus $\mathcal{C}$.\label{table:fit_new_vec-words}}
\end{table}
%%%%%%%%%%%%%%%%%%%%%%%%%%%%%%%%%%%%%%%%%%%%%%%%%%%%%%%%%%%%%%%%%%%%%%%%%%%%%%%%%%%%%%%%%

\section{Performance}

We evaluate the performance of \textsf{LEARN\_VECTOR} by considering the \emph{order} and the \emph{cosine score} of the learned vector $\hat{v}$ returned by the algorithm, defined below.

\subsection{Order and cosine score of the learned vector}\label{section:order-cosine_score}

Let $\hat{w}$ be one of the seven words we trained a vector for, $\hat{v}$ the learned vector returned by the algorithm, and $v_{\hat{w}} \in V_{\mathcal{D}}$ the ``true'' vector for word $\hat{w}$. Number the vectors $v_1, v_2, \ldots, v_{|\mathcal{D}|} \in V_{\mathcal{D}}$ in order of decreasing cosine similarity from $\hat{v}$, so that $v_1 \cdot \hat{v}  > v_2 \cdot \hat{v} > \cdots > v_{|\mathcal{D}|} \cdot \hat{v}$. Then the \emph{order} of $\hat{v}$ is the number $k \in \mathbb{N}$ such that $v_{\hat{w}} = v_k$, i.e., the true vector $v_{\hat{w}}$ has the $k$th largest cosine similarity from $\hat{v}$ out of all the words in $\mathcal{D}$. The \emph{cosine score} of $\hat{v}$ is the cosine similarity between the true vector $v_{\hat{w}}$ and the vector $\hat{v}$ returned by the algorithm, i.e., $v_{\hat{w}} \cdot \hat{v}$.

\subsection{Evaluation}

\begin{figure}
\includegraphics[width=\textwidth]{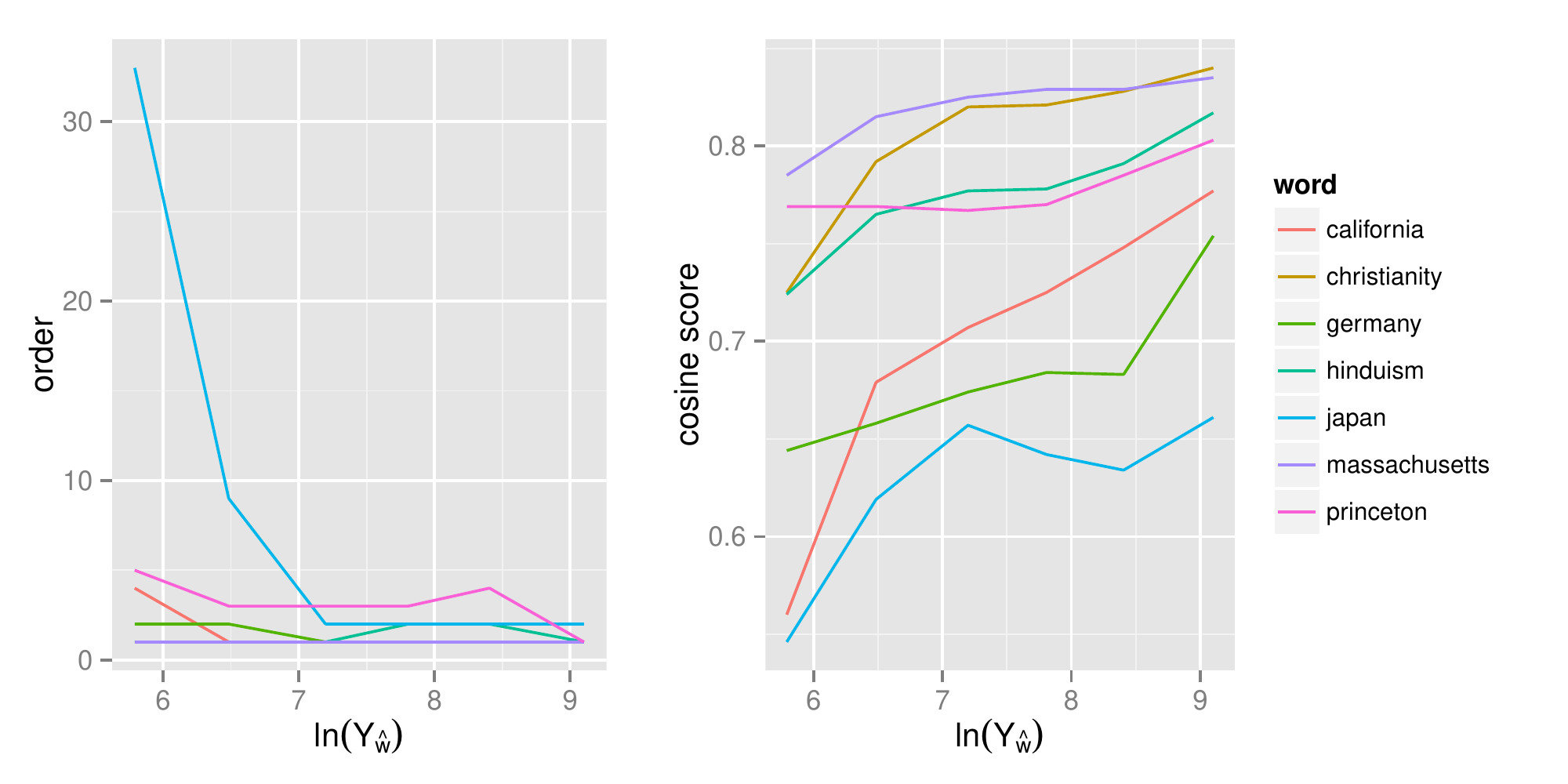}
\caption{The order and cosine score of the learned vector $\hat{v}$ returned by \textsf{LEARN\_VECTOR} (Algorithm \ref{section:fit_new_vec-algo}) for word $\hat{w}$, using varying amounts of co-occurrence data $Y_{\hat{w}}$. (See Section \ref{section:order-cosine_score} for the definitions of order and cosine score.) Note that in general, the order of $\hat{v}$ is decreasing, and the cosine score of $\hat{v}$ is increasing, in the amount of co-occurrence data $Y_{\hat{w}}$. (The occasional decrease in the cosine score is due to random initialization of the vector $\hat{v}$.) Also, observe that the words for which rank $k=10$ was used (\texttt{california}, \texttt{germany}, and \texttt{japan}) have lower cosine scores than the words for which rank $k=3$ was used. The algorithm achieves very good performance while using only a small fraction of the total amount of co-occurrence data in the Wikipedia corpus $\mathcal{C}$: For example, by using only $Y_{\hat{w}}(\Gamma)/X_{\hat{w}} = 1.846\textnormal{e}\!-\!02$ of the total amount of co-occurrence data for the word $w=\texttt{christianity}$, the algorithm is able to learn a vector whose order is 1 and cosine score is $0.84$. \label{fig:fit_new_vec}}
\end{figure}

In Figure \ref{fig:fit_new_vec}, we provide a plot of the order and cosine score of $\hat{v}$ for varying values of $\ln(Y_{\hat{w}})$, where $Y_{\hat{w}} := \sum_{w \in D'} Y_{\hat{w}w}$ is the amount of co-occurrence data for $\hat{w}$ in the training corpus $\Gamma$ with vocabulary set $D'$. Note that in general, the order of $\hat{v}$ is decreasing, and the cosine score of $\hat{v}$ is increasing, in the amount of co-occurrence data $Y_{\hat{w}}$. Moreover, the algorithm seems to achieve a higher cosine score by using a smaller rank $k$ for the basis of the category subspace: The words for which rank $k=10$ was used (\texttt{california}, \texttt{germany}, and \texttt{japan}) have lower cosine scores than the words for which rank $k=3$ was used.

We provide an explanation as to why using a smaller rank $k$ improves the algorithm's performance. For a category $c$, let $U_k$ be a rank-$k$ basis for the subspace of $c$. The regularization term $\lambda\len{v - U_k b}$ in the objective (\ref{eq:objective-fit_new_vec}) serves to train $\hat{v}$ \emph{near} the subspace $U_k$, which by definition only captures the general notion of category $c$. Recall from Chapter \ref{chapter:capture_rate} that the first basis vector $u_1$ of $U_k$ is the defining vector that encodes the most information about $c$, while the subsequent basis vectors $u_i$ for $i \geq 2$ capture more specific information about individual words in $c$. By using a smaller rank $k$, we throw away the more ``noisy'' vectors $u_i$ for $i \geq k \geq 1$, allowing $U_k$ to capture the generation notion of category $c$ better. This allows the regularization term to train the ``category'' component of $\hat{v}$ more accurately. Note that the other component, which is specific to word $\hat{w}$ and lies outside the category subspace, is trained by the term $g(Y_{ww'}) \paren*{||v +v_{w'}||^2 - \log(Y_{ww'}) +Z}^2$ in the objective (\ref{eq:objective-fit_new_vec}).

\begin{figure}
\begin{subfigure}{\textwidth}\centering
\includegraphics[width=\textwidth]{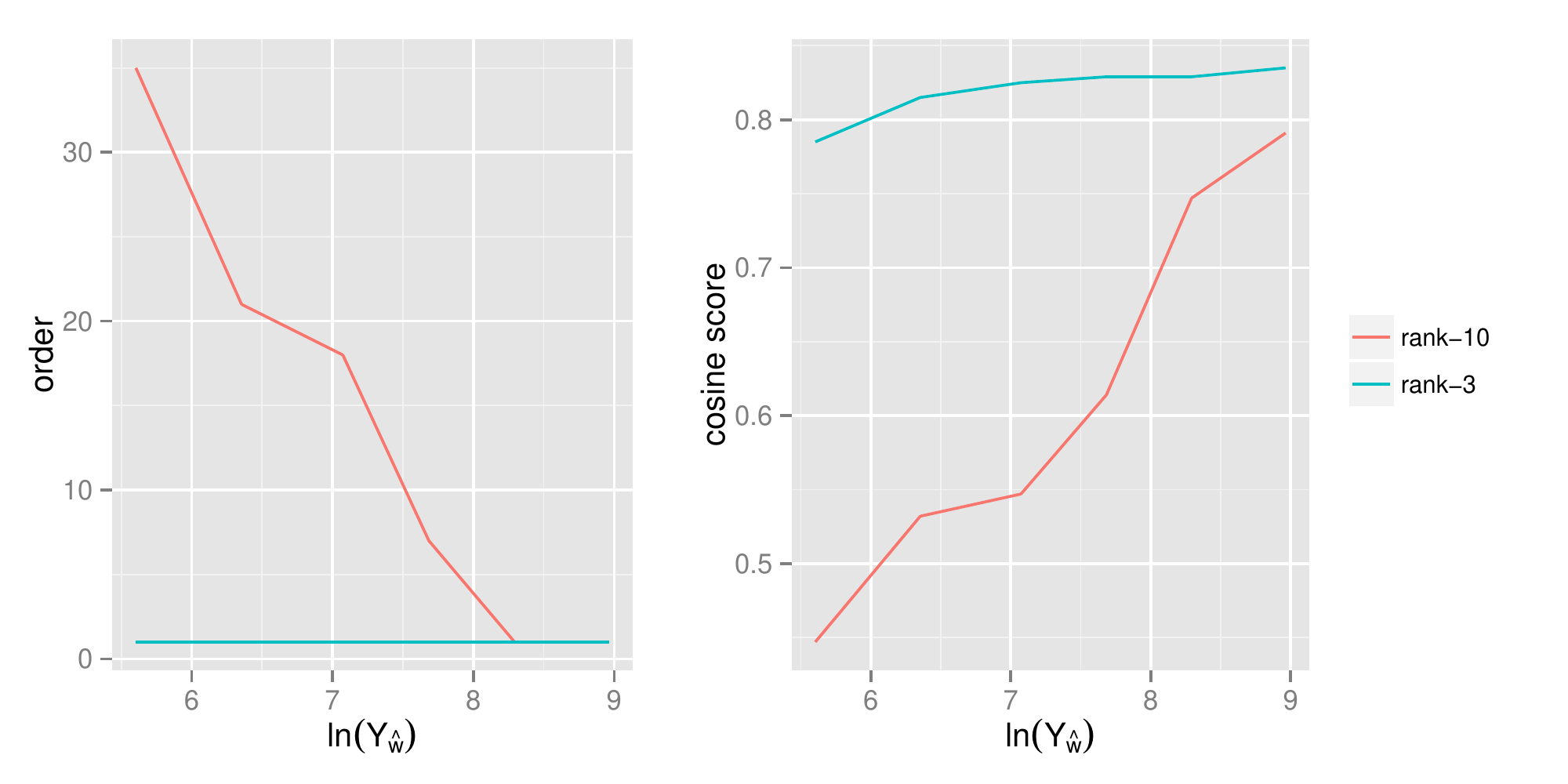}
\caption{$\hat{w}=\texttt{massachusetts}$}
\end{subfigure}

\begin{subfigure}{\textwidth}\centering
\includegraphics[width=\textwidth]{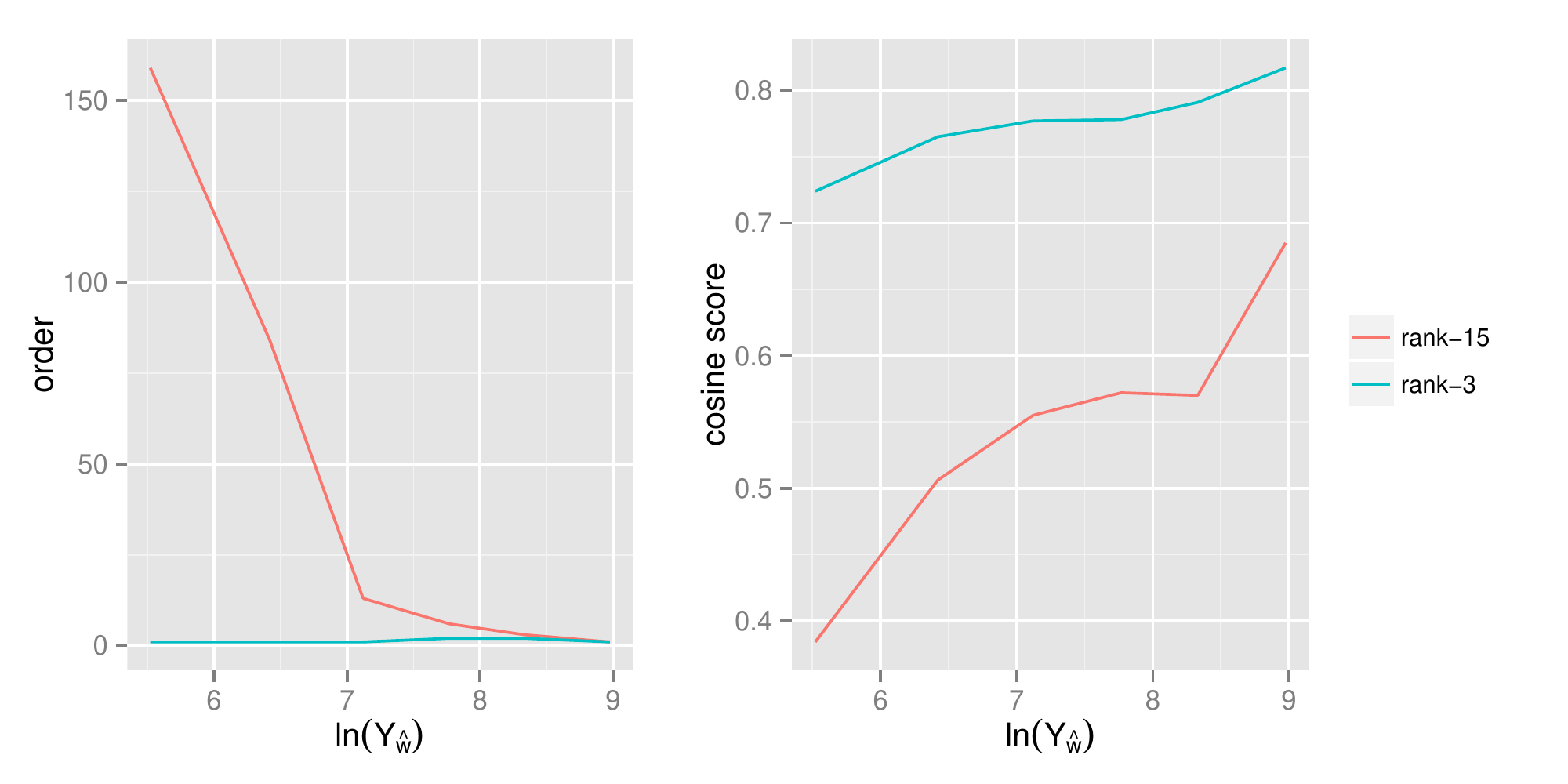}
\caption{$\hat{w}=\texttt{hinduism}$}
\end{subfigure}
\caption{Using \textsf{LEARN\_VECTOR} (Algorithm \ref{section:fit_new_vec-algo}), we trained two vectors for the same word $\hat{w}$, one using a low rank $k$ ({\color{blue}blue}) and the other using a high rank $k$ ({\color{red}red}), and compared their order and cosine score. (For each rank $k$, we tried various learning rates $\eta$ and regularization weights $\lambda$ to try to optimize performance; here, we provide the best performance found for each $k$.) For both $\hat{w}=\texttt{massachusetts}$ and $\hat{w}=\texttt{hinduism}$, the vector learned using the lower rank resulted in a lower order and a much higher cosine score. This demonstrates that using a smaller rank $k$ results in better performance for \textsf{LEARN\_VECTOR}.}
\label{fig:fit_new_vec-comparison}
\end{figure}

To illustrate our point, we trained two vectors for the same word $\hat{w}$, one using a low rank $k$ and the other using a high rank $k$, and compared their order and cosine score (see Figure \ref{fig:fit_new_vec-comparison}). For both $\hat{w}=\texttt{massachusetts}$ and $\hat{w}=\texttt{hinduism}$, the vector learned using the lower rank resulted in a lower order and a much higher cosine score. This demonstrates that using a smaller rank $k$ results in better performance for \textsf{LEARN\_VECTOR}.

To compare the amount of co-occurrence data for $\hat{w}$ in a subcorpus $\Gamma$ to the amount of co-occurrence data for $\hat{w}$ in the Wikipedia corpus $\mathcal{C}$, we look at the fraction $Y_{\hat{w}}(\Gamma)/X_{\hat{w}}$, which is listed in Table \ref{table:fit_new_vec-param}. Note that the algorithm achieves very good performance while using only a small fraction of the total amount of co-occurrence data in the Wikipedia corpus $\mathcal{C}$. For example, by using only $Y_{\hat{w}}(\Gamma)/X_{\hat{w}} = 1.846\textnormal{e}\!-\!02$ of the total amount of co-occurrence data for the word $w=\texttt{christianity}$, the algorithm is able to learn a vector whose order is 1 and cosine score is $0.84$.

Lastly, note that the performance depends heavily on the parameter values chosen, and can be further improved by fine-tuning the parameters.

\section{Conclusion}

We have demonstrated that, in principle, one can learn vectors with substantially less data by using the low-dimensional subspace of categories. An interesting experiment to try is the following: Use Algorithm \ref{section:fit_new_vec-algo} to learn vectors for rare words, and see if new facts can be discovered using these vectors. We leave this to future work.

Moreover, one can extend this method to learn vectors for any words -- even words that do not appear at all in the Wikipedia corpus -- using web-scraping tools, such as Google search, to obtain additional co-occurrence data. However, the corpora obtained from Google search may be drawn from a different distribution than the wikipedia corpus, and hence skew the data in a certain way. We leave this to future work.

One weakness of \textsf{LEARN\_VECTOR} is that it requires having prior knowledge of what category a word $\hat{w}$ belongs in. If our prior knowledge is wrong, then the fitted vector for $\hat{w}$ may be very bad. One could come up with an automatic method classifying which category $w$ belongs to.

%%%%%%%%%%%%%%%%%%%%%%%%%%%%%%%%%%%%%%%%%%%%%%%%%%%%%%%%%%%%%%%%%%%%%%%%%%%%%%%%%%%%%%%%%
\begin{table}\footnotesize\centering
\input{	table-fit_new_vec-param.tex}
\caption{For each word $\hat{w}$, we extracted six subcorpora $\Gamma_1, \ldots, \Gamma_6$ from the original corpus $\mathcal{C}$, where $\Gamma_1 \subset \Gamma_2 \subset \cdots \subset \Gamma_6$. We use $Y_{\hat{w}}(\Gamma) := \sum_{w \in D'} Y_{\hat{w}w}$ to quantify the amount of co-occurrence data for word $\hat{w}$ in a corpus $\Gamma$ with vocabulary set $D'$. For each $i \in \{1, \ldots, 6\}$, we trained the algorithm on the subcorpus $\Gamma_i$ for various learning rates $\eta_i$ and regularization weights $\lambda_i$. In the tables above, we list the parameter values $\eta_i$, $\lambda_i$ that resulted in the best performance (shown in Figure \ref{fig:fit_new_vec}). To compare the amount of co-occurrence data for $\hat{w}$ in $\Gamma$ to the amount of co-occurrence data for $\hat{w}$ in $\mathcal{C}$, we look at the fraction $Y_{\hat{w}}(\Gamma)/X_{\hat{w}}$. \label{table:fit_new_vec-param}}
\end{table}
%%%%%%%%%%%%%%%%%%%%%%%%%%%%%%%%%%%%%%%%%%%%%%%%%%%%%%%%%%%%%%%%%%%%%%%%%%%%%%%%%%%%%%%%%

%% file: table-fit_new_vec-param.tex
\begin{subtable}{.5\textwidth}\centering\subcaption{$w=\texttt{california}$}\begin{tabular}{ |c || c | c | c | c |}\hline $i$ & $\ln Y_w(\Gamma_i)$ & $Y_w(\Gamma_i)/X_w$ & $\eta_i$ & $\lambda_i$\\ \hline
$1$ & 5.459 & 5.735e-05 & 8.9e-05 & 1.72\\
$2$ & 6.353 & 1.403e-04 & 6.0e-05 & 1.70\\
$3$ & 7.087 & 2.921e-04 & 4.5e-05 & 1.70\\
$4$ & 7.688 & 5.332e-04 & 3.3e-05 & 1.67\\
$5$ & 8.297 & 9.802e-04 & 5.7e-05 & 0.81\\
$6$ & 8.993 & 1.966e-03 & 5.5e-05 & 0.60\\
\hline\end{tabular}\end{subtable}\begin{subtable}{.5\textwidth}\centering\subcaption{$w=\texttt{christianity}$}\begin{tabular}{ |c || c | c | c | c |}\hline $i$ & $\ln Y_w(\Gamma_i)$ & $Y_w(\Gamma_i)/X_w$ & $\eta_i$ & $\lambda_i$\\ \hline
$1$ & 5.353 & 5.064e-04 & 9.6e-05 & 2.27\\
$2$ & 6.408 & 1.453e-03 & 7.7e-05 & 3.14\\
$3$ & 7.090 & 2.876e-03 & 5.8e-05 & 3.14\\
$4$ & 7.681 & 5.191e-03 & 4.3e-05 & 3.14\\
$5$ & 8.311 & 9.746e-03 & 3.1e-05 & 2.96\\
$6$ & 8.949 & 1.846e-02 & 1.8e-05 & 2.75\\
\hline\end{tabular}\end{subtable}\vspace{5ex}
\begin{subtable}{.5\textwidth}\centering\subcaption{$w=\texttt{germany}$}\begin{tabular}{ |c || c | c | c | c |}\hline $i$ & $\ln Y_w(\Gamma_i)$ & $Y_w(\Gamma_i)/X_w$ & $\eta_i$ & $\lambda_i$\\ \hline
$1$ & 5.698 & 9.676e-05 & 8.9e-05 & 1.72\\
$2$ & 6.454 & 2.061e-04 & 1.0e-04 & 1.70\\
$3$ & 7.101 & 3.936e-04 & 6.0e-05 & 2.08\\
$4$ & 7.682 & 7.038e-04 & 3.3e-05 & 1.67\\
$5$ & 8.266 & 1.262e-03 & 2.3e-05 & 1.26\\
$6$ & 8.943 & 2.484e-03 & 4.9e-05 & 0.73\\
\hline\end{tabular}\end{subtable}\begin{subtable}{.5\textwidth}\centering\subcaption{$w=\texttt{hinduism}$}\begin{tabular}{ |c || c | c | c | c |}\hline $i$ & $\ln Y_w(\Gamma_i)$ & $Y_w(\Gamma_i)/X_w$ & $\eta_i$ & $\lambda_i$\\ \hline
$1$ & 5.521 & 2.409e-03 & 1.1e-04 & 2.89\\
$2$ & 6.419 & 5.917e-03 & 6.9e-05 & 3.14\\
$3$ & 7.120 & 1.193e-02 & 5.2e-05 & 3.14\\
$4$ & 7.768 & 2.279e-02 & 4.3e-05 & 3.14\\
$5$ & 8.330 & 3.999e-02 & 3.1e-05 & 2.96\\
$6$ & 8.976 & 7.627e-02 & 4.7e-05 & 2.31\\
\hline\end{tabular}\end{subtable}\vspace{5ex}
\begin{subtable}{.5\textwidth}\centering\subcaption{$w=\texttt{japan}$}\begin{tabular}{ |c || c | c | c | c |}\hline $i$ & $\ln Y_w(\Gamma_i)$ & $Y_w(\Gamma_i)/X_w$ & $\eta_i$ & $\lambda_i$\\ \hline
$1$ & 5.531 & 9.252e-05 & 8.9e-05 & 1.72\\
$2$ & 6.481 & 2.392e-04 & 6.0e-05 & 1.70\\
$3$ & 7.110 & 4.485e-04 & 6.0e-05 & 2.08\\
$4$ & 7.693 & 8.036e-04 & 3.3e-05 & 1.67\\
$5$ & 8.276 & 1.440e-03 & 2.3e-05 & 1.26\\
$6$ & 8.946 & 2.813e-03 & 4.3e-05 & 0.73\\
\hline\end{tabular}\end{subtable}\begin{subtable}{.5\textwidth}\centering\subcaption{$w=\texttt{massachusetts}$}\begin{tabular}{ |c || c | c | c | c |}\hline $i$ & $\ln Y_w(\Gamma_i)$ & $Y_w(\Gamma_i)/X_w$ & $\eta_i$ & $\lambda_i$\\ \hline
$1$ & 5.604 & 2.069e-04 & 1.1e-03 & 2.95\\
$2$ & 6.354 & 4.380e-04 & 9.2e-05 & 4.40\\
$3$ & 7.073 & 8.994e-04 & 7.0e-05 & 4.40\\
$4$ & 7.684 & 1.656e-03 & 4.7e-05 & 4.08\\
$5$ & 8.293 & 3.045e-03 & 3.4e-05 & 3.85\\
$6$ & 8.964 & 5.960e-03 & 2.2e-05 & 3.25\\
\hline\end{tabular}\end{subtable}\vspace{5ex}
\begin{subtable}{.5\textwidth}\centering\subcaption{$w=\texttt{princeton}$}\begin{tabular}{ |c || c | c | c | c |}\hline $i$ & $\ln Y_w(\Gamma_i)$ & $Y_w(\Gamma_i)/X_w$ & $\eta_i$ & $\lambda_i$\\ \hline
$1$ & 5.791 & 1.157e-03 & 1.3e-04 & 2.89\\
$2$ & 6.484 & 2.314e-03 & 1.0e-04 & 4.00\\
$3$ & 7.197 & 4.723e-03 & 7.8e-05 & 4.00\\
$4$ & 7.807 & 8.689e-03 & 5.8e-05 & 4.00\\
$5$ & 8.405 & 1.580e-02 & 4.1e-05 & 3.77\\
$6$ & 9.101 & 3.170e-02 & 6.2e-05 & 2.94\\
\hline\end{tabular}\end{subtable}

%% file: wordnet_filter.tex
\chapter{Using an external knowledge source to reduce false-positive rate}\label{chapter:wordnet_filter}
\chaptermark{Using external knowledge}
One can use external knowledge sources such as a dictionary or Wordnet \cite{wn} to \emph{filter} false-positive answers and improve accuracy on analogy queries. In this section, we focus on analogy queries ``a:b::c:??'' where there exists categories $c_A$, $c_B$ such that both $a$ and $c$ belong in $c_A$, and both $b$ and the correct answer $d$ belong in $c_B$.  In other words, $(a,b)$ and $(c,d)$ both satisfy a common relation $r$ that is well-defined.

\textsf{SOLVE\_QUERY} (Algorithm \ref{section:analogy_query-algo}) is a generic method for returning the top $N$ answers to an analogy query ``a:b::c:??''.\footnote{It is also used in other works (e.g. \cite{MSC+13, GloVe, arora}) to evaluate a method's performance on analogy tasks.} In Section \ref{section:wordnet}, we present two ways for filtering the candidate list of answers to reduce false-positive rate: \emph{POS filter} and \emph{LEX filter}. Figures \ref{fig:wordnet_filter-facts}, \ref{fig:wordnet_filter-gram}, and \ref{fig:wordnet_filter-semantics} compare the accuracy of \textsf{SOLVE\_QUERY} with and without these filters on 24 different relations $r$. We show that the POS filter always \emph{increases} the accuracy (either slightly or significantly, depending on the relation $r$), unless the accuracy is already $100\%$ without filter. On the other hand, the performance for the LEX filter varies widely, depending on the nature of the relation $r$. When used on appropriate relations $r$, such as facts-based relations (see Figure \ref{fig:freebase_relation}(a)-(h)), the LEX filter can improve the accuracy by as much as $+19.2\%$ than without filter, and $+17.7\%$ than with POS filter (see Figure \ref{fig:wordnet_filter-facts}(a)).

\section{Analogy queries}

Recall that word embeddings allow one to solve analogy queries of the form ``a:b::c:??'' using simple vector arithmetics. More specifically, for two word pairs $(a,b), (c,d)$ satisfying a common relation $r$, their word vectors satisfy
\[
v_a - v_b \approx v_c - v_d.
\]
Hence, a method to solve the analogy query ``a:b::c:??'' is to find the word $d \in \mathcal{D}$ whose vector $v_d$ is closest to $v_b - v_a + v_c$.

Given a set of words $\Delta \subseteq \mathcal{D}$ and a number $N \in \{1, \ldots, |\Delta|\}$, \textsf{SOLVE\_QUERY} (Algorithm \ref{section:analogy_query-algo}) returns the top $N$ answers in $\Delta$ for the analogy query ``a:b::c:??''. More specifically, it returns $N$ words in $\Delta$ corresponding to the top $N$ vectors in $V_{\Delta} := \{ v_w : w \in \Delta \}$ that are closest to the vector $v_b - v_a + v_c$. We say \textsf{SOLVE\_QUERY}  \emph{returns the correct answer} if the correct answer $d$ is in the set returned by the algorithm.

%%%%%%%%%%%%%%%%%%%%%%%%%%%%%%%%%%%%%%%%%%%%%%%%%%%%%%%%%%%%%%%%%%%%%%%%%%%%%%%%
\renewcommand{\figurename}{Algorithm}
\begin{figure}
\begin{framed}
\noindent {\bf function $\textsf{SOLVE\_QUERY}(\{a,b,c\}, \Delta, N)$}: Returns a list of $N$ words in $\Delta$.\\\\
\noindent Inputs:
\begin{itemize}
\item Words $a,b,c$ for which we want to solve the analogy query ``a:b::c:??''
\item $\Delta \subseteq \mathcal{D}$, a set of candidate answers
\item $N \in \{1, 2, \ldots, |\Delta| \}$, the number of answers to return.
\end{itemize}

\begin{enumerate}[Step 1.]
\item Let $V_{\Delta} := \{ v_w : w \in \Delta \}$.  Number the vectors $v_1, v_2, \ldots, v_{|\Delta|} \in V_{\Delta}$ in order of decreasing cosine similarity from the vector $v_b - v_a + v_c$. Let $S :=\{ v_1, v_2, \ldots, v_N \}$.
\item Return the list $\{ w \in \Delta : v_w \in S \}$.
\end{enumerate}

\noindent {\bf end function}
\end{framed}
\caption{$\textsf{SOLVE\_QUERY}(\{a,b,c\}, \Delta, N)$ returns top $N$ answers from $\Delta$ for the analogy query ``a:b::c:??''.}
\label{section:analogy_query-algo}
\end{figure}
\renewcommand{\figurename}{Figure}
%%%%%%%%%%%%%%%%%%%%%%%%%%%%%%%%%%%%%%%%%%%%%%%%%%%%%%%%%%%%%%%%%%%%%%%%%%%%%%%%

%%%%%%%%%%%%%%%%%%%%%%%%%%%%
\section{Wordnet}\label{section:wordnet}

In Wordnet, each word is labeled with POS (``part-of-speech'') and LEX (``lexicographic'') tags. The POS tag indicates the syntactic category of a word, such as \texttt{noun}, \texttt{verb}, \texttt{adjective}, and \texttt{adverb}. The LEX tag is more specific: See Table \ref{table:lexnames} for a complete list of the LEX tags in Wordnet. For any word $w \in \mathcal{D}$, let $\pos(w)$ and $\lex(w)$ be the set of POS and LEX tags for $w$, respectively. For example, for the currency word $w=\texttt{euro}$,
\begin{align*}
\pos(\texttt{euro}) &= \{ \texttt{noun} \},\\
\lex(\texttt{euro}) &= \{ \texttt{noun.quantity} \}.
\end{align*}

Define the following sets:
\begin{align*}
\mathcal{D}_{\pos(w)} &:= \{ w' \in \mathcal{D} : \pos(w') \cap \pos(w) \neq \emptyset \},\\
\mathcal{D}_{\lex(w)} &:= \{ w' \in \mathcal{D} : \lex(w') \cap \lex(w) \neq \emptyset \}.
\end{align*}
In other words, $\mathcal{D}_{\pos(w)}$ is the set of words that share a common POS tag with $w$, and similarly, $\mathcal{D}_{\lex(w)}$ is the set of words that share a common LEX tag with $w$. For example, $\mathcal{D}_{\pos(\texttt{euro})}$ contains all the \texttt{noun} words in $\mathcal{D}$, and  $\mathcal{D}_{\lex(\texttt{euro})}$ contains words such as \texttt{dollar} and \texttt{kilometer} which have the LEX tag $\texttt{noun.quantity}$.

Consider the analogy query ``a:b::c:??'' where there exists categories $c_A$ and $c_B$ such that both $a$ and $c$ belong in $c_A$, and both $b$ and the correct answer $d$ belong in $c_B$. If we assume that every word in category $c_B$ share a common POS (or LEX) tag, then we can use Wordnet to filter out words in $\mathcal{D}$ which cannot belong in $c_B$. More specifically, we only search among the words in $\mathcal{D}_{\pos(b)}$ (or $\mathcal{D}_{\lex(b)}$) for the correct answer $d$. Note that LEX is a \emph{stronger} filter than POS, in the sense that $\mathcal{D}_{\lex(b)} \subset \mathcal{D}_{\pos(b)}$. 

\begin{table}\centering
\input{lexnames.tex}
\caption{List of all LEX tags in Wordnet.\label{table:lexnames}}
\end{table}

\section{Experiment}

We tested \textsf{SOLVE\_QUERY} (Algorithm \ref{section:analogy_query-algo}) on 24 relations from Figure \ref{fig:freebase_relation} in the following manner. For each relation $r$, let $S_r \subset \mathcal{D}_r$ be the set of word pairs contained in the correponding relation file (see Figure \ref{fig:freebase_relation}). For each $N \in \{1, 5, 10, 25, 50\}$, we performed the following experiment:
\begin{enumerate}[Step 1.]

\item Initialize $n = n_{\pos} = n_{\lex} = 0$.
\item For each $(a,b) \in S_r$, and for each $(c,d) \in S_r$ such that $(c,d) \neq (a,b)$, solve the analogy query ``a:b::c:??'' using Algorithm \ref{section:analogy_query-algo}:
\begin{align*}
S &\leftarrow \textsf{SOLVE\_QUERY}(\{a,b,c\}, \mathcal{D}, N) \\
S_{\pos} &\leftarrow \textsf{SOLVE\_QUERY}(\{a,b,c\}, \mathcal{D}_{\pos(b)}, N) \\
S_{\lex} &\leftarrow\textsf{SOLVE\_QUERY}(\{a,b,c\}, \mathcal{D}_{\lex(b)}, N).
\end{align*}
We say $S$, $S_{\pos}$, and $S_{\lex}$ are the answers returned by the algorithm \emph{without filter}, \emph{with POS filter}, and \emph{with LEX filter}, respectively. 
\begin{itemize}
\item If $d \in S$, then increment $n$ by 1.
\item If $d \in S_{\pos}$, then increment $n_{\pos}$ by 1.
\item If $d \in S_{\lex}$, then increment $n_{\lex}$ by 1.
\end{itemize}
\end{enumerate}

In other words, we test the algorithm on the analogy queries ``a:b::c:??'' and ``c:d::a:??'' for every possible pairs $(a,b), (c,d) \in S_r$. The total number of times the algorithm returns the correct answer without filter, with POS filter, and with LEX filter are $n$, $n_{\pos}$, and $n_{\lex}$, respectively. Since the total number of analogy queries tested on is $|S_r|(|S_r|-1)$, the \emph{accuracy} of the algorithm without filter, with POS filter, and with LEX filter are given by $\frac{n}{|S_r|(|S_r|-1)}$, $\frac{n_{\pos}}{|S_r|(|S_r|-1)}$, and  $\frac{n_{\lex}}{|S_r|(|S_r|-1)}$, respectively.

\section{Performance}

\begin{center}\begin{figure}
\includegraphics[width=\textwidth]{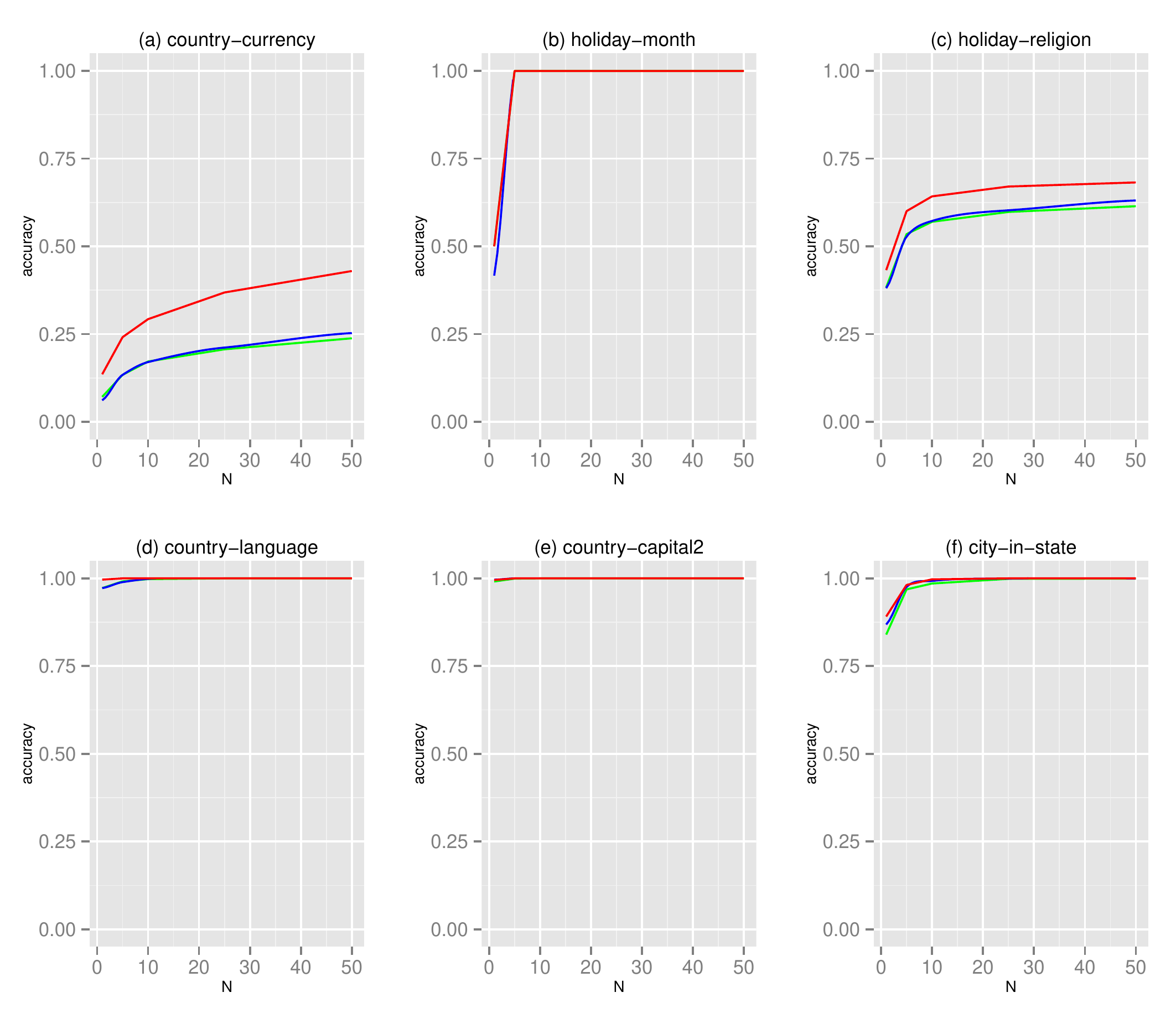}
\caption{Accuracy of the algorithm without filter ({\color{green}green}), with POS filter ({\color{blue}blue}), and with LEX filter ({\color{red}red}) on the facts-based relation files from Figure \ref{fig:freebase_relation}(a)-(f). For (a) and (c), the LEX filter improves the accuracy significantly, while the POS filter improves the accuracy only slightly. For all other relation files, the algorithm already achieves an accuracy of 1 without filter.\label{fig:wordnet_filter-facts}}
\end{figure}\end{center}

\begin{center}\begin{figure}
\includegraphics[width=\textwidth]{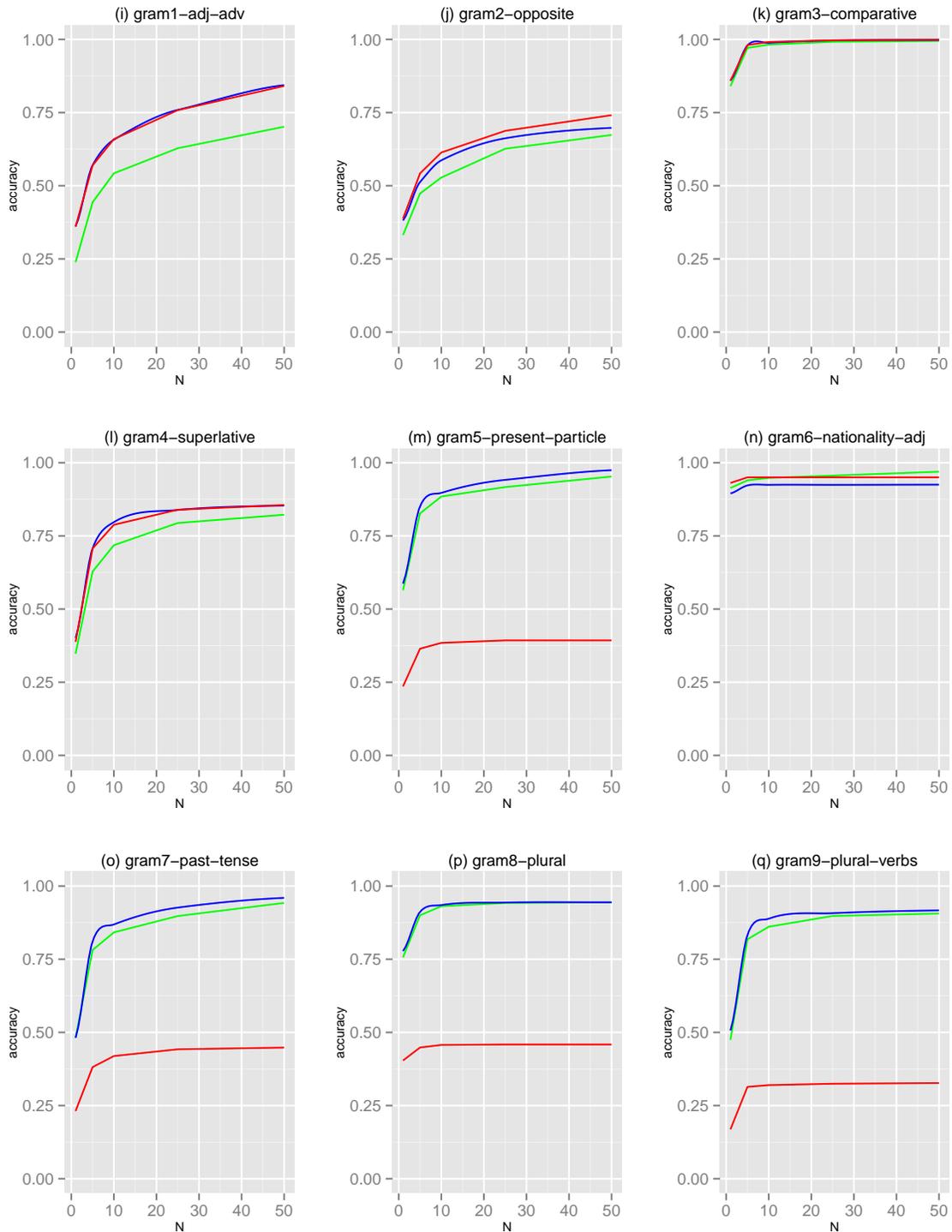}
\caption{Accuracy of the algorithm without filter ({\color{green}green}), with POS filter ({\color{blue}blue}), and with LEX filter ({\color{red}red}) on the grammar-based relation files from Figure \ref{fig:freebase_relation}(i)-(q). For (n), both the LEX filter and the POS filter worsen the accuracy slightly, due to the fact that Wordnet does not have an entry for the word ``argentinean''. For all other relation files, POS improves the accuracy by a modest amount. LEX improves the accuracy for (i), (j), and (l), but performs very poorly for (m), (o)-(q), due to the fact that words belonging to $c_B$ have different LEX tags, causing LEX to filter out the correct answer. \label{fig:wordnet_filter-gram}}
\end{figure}\end{center}

\begin{center}\begin{figure}
\includegraphics[width=\textwidth]{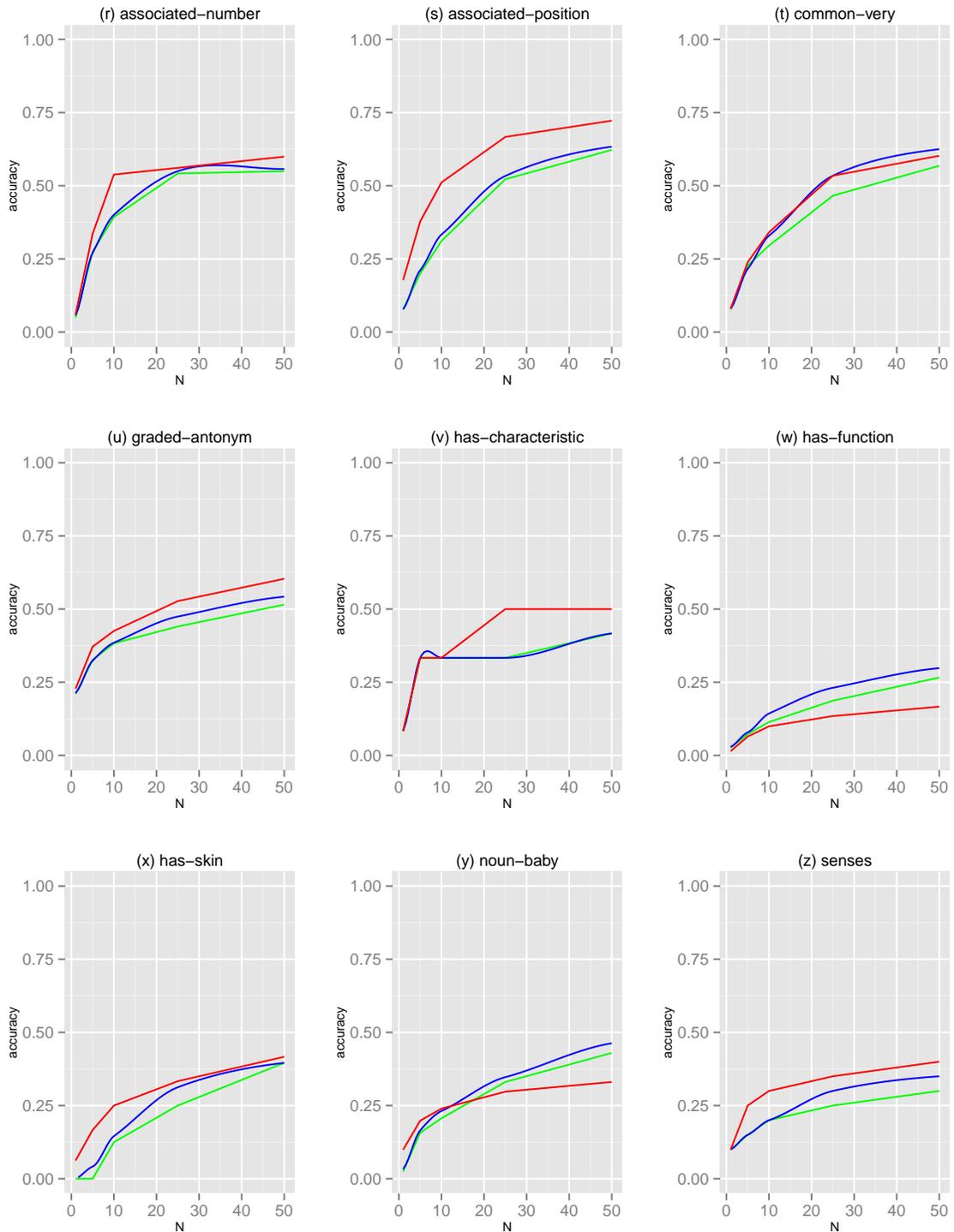}
\caption{Accuracy of the algorithm without filter ({\color{green}green}), with POS filter ({\color{blue}blue}), and with LEX filter ({\color{red}red}) on the semantics-based relation files from Figure \ref{fig:freebase_relation}(r)-(z). LEX filter worsens the accuracy for (w) and (y), but improves the accuracy significantly on all other relation files. POS filter consistently improves the accuracy on all relation files. \label{fig:wordnet_filter-semantics}}
\end{figure}\end{center}

Figures \ref{fig:wordnet_filter-facts}, \ref{fig:wordnet_filter-gram}, and \ref{fig:wordnet_filter-semantics} compare the accuracy of \textsf{SOLVE\_QUERY} with and without filters on 24 different relations $r$, which are taken from Figure \ref{fig:freebase_relation}. We ignore the performance on relation (n) \texttt{gram6-nationality-adj} in Figure \ref{fig:wordnet_filter-gram}, due to the fact that Wordnet does not have an entry for the word ``argentinean'' which is included in the test bed.\footnote{So for all analogy queries ``a:b::c:??'' where  either $b$ or the correct answer $d$ is the word \texttt{argentinean}, both POS and LEX filter out the correct answer from $\mathcal{D}$, causing the algorithm to get the analogy query wrong and therefore lower its accuracy slightly.}

Depending on the relation $r$, the POS filter always \emph{increases} the accuracy, either slightly (see (a), (c), (j), (l), (m), (o)-(z)) or significantly (see (i)), unless the accuracy is already $100\%$ without filter (see (b), (d), (e), (f), (k)).

On the other hand, the performance for the LEX filter varies widely depending on the nature of the relation $r$, due to the fact that LEX is a stronger filter than POS. For relations where words belonging to $c_B$ share a common LEX tag (e.g., (a), (c), (i), (j), (l), (r)-(v), (z)), LEX improves the accuracy significantly by filtering out false-positive answers. On the contrary, for relations where words belonging to $c_B$ have different LEX tags (e.g., (m), (o)-(q), (w), (y)), LEX filters out the correct answer from $\mathcal{D}$, and hence worsens the accuracy significantly. When used on appropriate relations $r$, such as facts-based relations (see Figure \ref{fig:freebase_relation}(a)-(h)), the LEX filter can improve the accuracy by as much as $+19.2\%$ than without filter, and $+17.7\%$ than with POS filter (see the plot in Figure \ref{fig:wordnet_filter-facts}(a) for $N=50$).

\section{Conclusion}

We have shown that external knowledge sources such as Wordnet can be used to improve accuracy on analogy queries, sometimes significantly, by filtering out false-positive answers. As an extension of the idea, we can apply the Wordnet filter to \textsf{EXTEND\_CATEGORY} (Algorithm \ref{section:discover_new_entities-algo}) for learning new words in a category, or \textsf{EXTEND\_RELATION} (Algorithm \ref{section:learn_new_relations-algo}) for learning new word pairs in a relation, to decrease the false-positive rate and improve its performance. We leave this to future work.

%% file: lexnames.tex
\begin{tabular}{| l | t | l |}
\hline
\# & LEX tag & Description \\ \hline
00 & adj.all & all adjective clusters\\
01 & adj.pert & relational adjectives (pertainyms)\\
02 & adv.all & all adverbs\\
03 & noun.Tops & unique beginner for nouns\\
04 & noun.act & nouns denoting acts or actions\\
05 & noun.animal & nouns denoting animals\\
06 & noun.artifact & nouns denoting man-made objects\\
07 & noun.attribute & nouns denoting attributes of people and objects\\
08 & noun.body & nouns denoting body parts\\
09 & noun.cognition & nouns denoting cognitive processes and contents\\
10 & noun.communication & nouns denoting communicative processes and contents\\
11 & noun.event & nouns denoting natural events\\
12 & noun.feeling & nouns denoting feelings and emotions\\
13 & noun.food & nouns denoting foods and drinks\\
14 & noun.group & nouns denoting groupings of people or objects\\
15 & noun.location & nouns denoting spatial position\\
16 & noun.motive & nouns denoting goals\\
17 & noun.object & nouns denoting natural objects (not man-made)\\
18 & noun.person & nouns denoting people\\
19 & noun.phenomenon & nouns denoting natural phenomena\\
20 & noun.plant & nouns denoting plants\\
21 & noun.possession & nouns denoting possession and transfer of possession\\
22 & noun.process & nouns denoting natural processes\\
23 & noun.quantity & nouns denoting quantities and units of measure\\
24 & noun.relation & nouns denoting relations between people or things or ideas\\
25 & noun.shape & nouns denoting two and three dimensional shapes\\
26 & noun.state & nouns denoting stable states of affairs\\
27 & noun.substance & nouns denoting substances\\
28 & noun.time & nouns denoting time and temporal relations\\
29 & verb.body & verbs of grooming, dressing and bodily care\\
30 & verb.change & verbs of size, temperature change, intensifying, etc.\\
31 & verb.cognition & verbs of thinking, judging, analyzing, doubting\\
32 & verb.communication & verbs of telling, asking, ordering, singing\\
33 & verb.competition & verbs of fighting, athletic activities\\
34 & verb.consumption & verbs of eating and drinking\\
35 & verb.contact & verbs of touching, hitting, tying, digging\\
36 & verb.creation & verbs of sewing, baking, painting, performing\\
37 & verb.emotion & verbs of feeling\\
38 & verb.motion & verbs of walking, flying, swimming\\
39 & verb.perception & verbs of seeing, hearing, feeling\\
40 & verb.possession & verbs of buying, selling, owning\\
41 & verb.social & verbs of political and social activities and events\\
42 & verb.stative & verbs of being, having, spatial relations\\
43 & verb.weather & verbs of raining, snowing, thawing, thundering\\
44 & adj.ppl & participial adjectives\\
\hline
\end{tabular}

%% file: conclusion.tex
\chapter{Conclusion}

We have demonstrated that the linear algebraic structure of word embeddings can be used to reduce data requirements for methods of learning facts. In particular, we demonstrated that categories and relations form a low-rank subspace $U_k = \{ u_1, \ldots, u_k \}$ in the projected space (Chapter \ref{chapter:capture_rate}), and this subspace can be used to discover new facts with fairly low false-positive rates (\ref{chapter:learn_new_facts}) and learn new vectors for words with substantially less co-occurrence data (Chapter \ref{chapter:fit_new_vec}).

In Chapter \ref{chapter:capture_rate}, we demonstrated that the first basis vector $u_1$ of a low-rank subspace encodes the most \emph{general} information about a category $c$ (or a relation $r$), whereas the subsequent basis vectors $u_i$ for $i \geq 2$ encode more ``specific'' information pertaining to individual words in $\mathcal{D}_c$ (or word pairs in $\mathcal{D}_r$). It remains to be discovered what specific features are captured by these basis vectors for various categories and relations. For example, if $U_k$ is a basis for the subspace of category $c=\texttt{country}$, then perhaps having a positive second coordinate $v_w \cdot u_2$ in the basis indicates that $w$ is a \emph{developed} country, and having a negative fourth coordinate $v_w \cdot u_4$ indicates that country $w$ is located in Europe. We leave this to future work.

In Chapter \ref{chapter:learn_new_facts}, we used the low-dimensional subspace of categories and relations to discover new facts with fairly low false-positive rates. The performance of \textsf{EXTEND\_RELATION} (Algorithm \ref{section:learn_new_relations-algo}) is fairly surprising, given the simplicity of the algorithm and the fact that it returns plausible word pairs out of all possible word pairs in $\mathcal{D} \times \mathcal{D}$. At the very least, \textsf{EXTEND\_RELATION} has shown to drastically narrow down the search space for discovering new word pairs satisfying a given relation, and allows for sharper classification than simple clustering. It can supplement other methods for extending knowledge bases to improve efficiency and attain even higher accuracy rates. One could also combine \textsf{EXTEND\_RELATION} with POS and LEX filters described in Chapter \ref{chapter:wordnet_filter}, or explore other ways of utilizing external knowledge sources (e.g., a dictionary) to filter false-positive answers.

In Chapter \ref{chapter:fit_new_vec} we demonstrated that, in principle, one can learn vectors with substantially less data by using the low-dimensional subspace of categories. An interesting experiment to try is the following: Use \textsf{LEARN\_VECTOR}  to learn vectors for rare words in the Wikipedia corpus, and see if new facts can be discovered using these vectors. It would also be interesting to try variants of the objective (\ref{eq:objective-fit_new_vec}), perhaps by adding the regularization term $\lambda \len{\hat{v} - U_kb}^2$ to other existing objectives such as GloVe \cite{GloVe}.

Moreover, one can extend this method to learn vectors for \emph{any} words -- even words that do not appear at all in the Wikipedia corpus -- using web-scraping tools, such as Google search, to obtain additional co-occurrence data. However, the corpora obtained from Google search may be drawn from a different distribution than the Wikipedia corpus, and hence could skew the data in a certain way. We leave this to future work.